\documentclass[10pt,twocolumn,letterpaper]{article}

\usepackage[accsupp]{axessibility}
\usepackage{cvpr}              
\usepackage[utf8]{inputenc} 
\usepackage[T1]{fontenc}    
\usepackage{hyperref}       
\usepackage{url}            
\usepackage{booktabs}       
\usepackage{amsfonts}       
\usepackage{nicefrac}       
\usepackage{microtype}      
\usepackage{wrapfig}
\usepackage{adjustbox}
\usepackage{amsmath}
\usepackage{graphicx}
\usepackage[accsupp]{axessibility}

\usepackage{physics}
\usepackage{multirow}

\usepackage[table,xcdraw]{xcolor}

\usepackage{makecell}

\usepackage[most]{tcolorbox}
\tcbset{
    sharp corners,
    boxrule=0.8pt,
    colback=green!5!white, 
    colframe=teal!80!black, 
    arc=3mm, 
    drop shadow, 
    boxsep=5pt, 
    left=5pt, right=5pt, top=5pt, bottom=5pt 
}

\def\paperID{10604} 
\def\confName{CVPR}
\def\confYear{2026}

\title{Revisiting Model Inversion Evaluation: \\ From Misleading Standards to Reliable Privacy Assessment}

\author{%
  Sy-Tuyen Ho$^\diamondsuit$$^\clubsuit$\thanks{\* Equal contribution. Work done while at SUTD. Accepted to CVPR 2026 Findings} \hspace{5 mm} Koh Jun Hao$^\diamondsuit$$^*$ \hspace{5 mm} Ngoc-Bao Nguyen$^\diamondsuit$$^*$ \\ {Alexander Binder}$^\heartsuit$ \hspace{5 mm} {Ngai-Man Cheung}$^\diamondsuit$ \\
$^\diamondsuit$ Singapore University of Technology and Design (SUTD) \\
$^\clubsuit$ University of Maryland College Park \\
$^\heartsuit$ Leipzig University \\
\texttt{ngaiman\_cheung@sutd.edu.sg} \\
\texttt{stho@umd.edu}
}

\begin{document}
\maketitle
\begin{abstract}

Model Inversion (MI) attacks aim to reconstruct information from private training data by exploiting access to a target model. Nearly all recent MI studies evaluate attack success using a standard framework that computes attack accuracy through a secondary evaluation model trained on the same private data and task design as the target model. In this paper, we present the first in-depth analysis of this dominant evaluation framework and reveal a fundamental issue: many reconstructions deemed ``successful'' under the existing framework are in fact false positives that do not capture the visual identity of the target individual. As our first contribution, we show that
\textbf{these MI false positives satisfy the same formal conditions as Type I adversarial examples}. In addition, through controlled experiments, we demonstrate extremely high false-positive transferability, an empirical signature characteristic of adversarial behavior, indicating that many MI false positives likely contain Type I adversarial features. This adversarial transferability significantly inflates reported attack accuracy and leads to an overstatement of privacy leakage in existing MI work. To address this issue, as our second contribution, we introduce \textbf{a new evaluation framework based on Multimodal Large Language Models (MLLMs)}, whose general-purpose visual reasoning avoids the shared-task vulnerability and reduces Type-I adversarial transferability of current evaluation framework. We propose systematic design principles for MLLM-based evaluation. Using this framework, we reassess 27 MI attack setups across diverse datasets, target models, and priors, and find consistently high false-positive rates under the conventional approach. Our results call for a reevaluation of progress in MI research and establish MLLM-based evaluation as a more reliable standard for assessing privacy risks in machine learning systems. \textbf{Code/data/prompt are available at \url{https://hosytuyen.github.io/projects/FMLLM}}

\end{abstract}    
\section{Introduction}
\label{sec:intro}

Model Inversion (MI) attacks pose a significant privacy threat by attempting to reconstruct confidential information from sensitive training data through exploiting access to machine learning models. 
Recent state-of-the-art (SOTA) MI attacks \citep{zhang2020secret, chen2021knowledge, wang2021variational, nguyen_2023_CVPR, nguyen2023labelonly, qiu2024closer, yuan2023pseudo, peng2024pseudoprivate, li2025head} have shown considerable advancements, reporting attack success rates exceeding 90\%. This vulnerability is particularly alarming for security-sensitive applications such as face recognition 
\citep{meng2021magface, guo2020learning, huang2020curricularface, schroff2015facenet}, medical diagnosis \citep{dufumier2021contrastive, yang2022towards, dippel2021towards}, or speech recognition \citep{chang2020end, krishna2019speech}.
The model under an MI attack is referred to as the {\em target model} $T$.

\begin{figure*}[t]
   
\vspace{-0.3cm}
  \centering
  \includegraphics[width=0.85\textwidth]{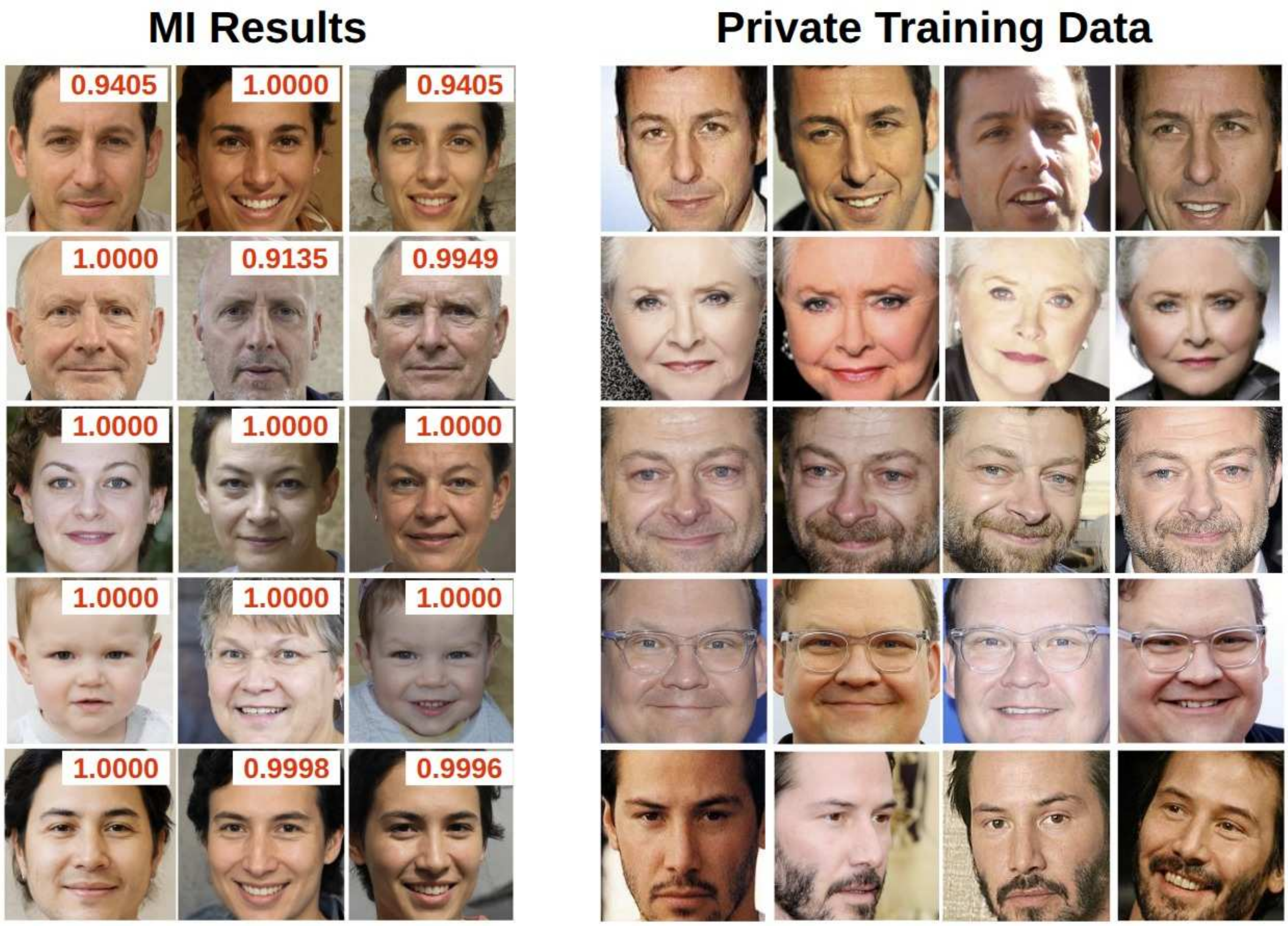}
  \vspace{-0.3cm}
  \caption{\textbf{We present the first and in-depth study on the Model Inversion (MI) evaluation}. Particularly, we investigate the most common MI evaluation framework $\mathcal{F}_{Curr}$ to measure MI Attack Accuracy (AttAcc). $\mathcal{F}_{Curr}$ is introduced in \citep{zhang2020secret} and is utilized to assess almost all recent MI attacks/defenses. However, we find that  $\mathcal{F}_{Curr}$ suffers from a significant number of false positives. These false positive MI reconstructed samples do not capture visual identity features of the target individual in the private training data, but they are still deemed successful attacks according to $\mathcal{F}_{Curr}$ with a high confidence (indicated in red text). Extensive false positives are in the Supp. }
  \label{fig:Teaser}
  \vspace{-0.4cm}

\end{figure*}

\noindent\textbf{Research gap.} Recently, there are many studies on improving MI attacks \citep{zhang2020secret,fredrikson2014privacy,an2022mirror,chen2021knowledge,yuan2023pseudo,qiu2024closer,nguyen2023labelonly,nguyen_2023_CVPR,kahla2022label,han2023reinforcement} and MI defenses \citep{wang2021improving,peng2022bilateral,struppek2024be,ho2024model,koh2024vulnerability}. To assess the effectiveness of these MI attacks/defenses, MI Attack Accuracy (AttAcc) is a standard and the most important metric. To measure AttAcc, almost all recent MI studies adopt the evaluation framework introduced by \citep{zhang2020secret}, which we denote as $\mathcal{F}_{Curr}$. Under $\mathcal{F}_{Curr}$, an {\em evaluation model} $E$ is used to predict the identities of individuals based on the MI-reconstructed images. Model $E$ is trained on {\em the  same private dataset and follows the same training task design} as in target model $T$. 
In particular, consider the scenario where the adversary targets a specific identity $y$. From this point onward, we use 
$y$ exclusively to denote the target class (label) in the MI attack, rather than a label predicted by any model.
The adversary produces reconstructed images  $x^{r}_{y}$ of the target label $y$ by exploiting access to the target model $T$. To measure AttAcc, these $x^{r}_{y}$ samples are then passed through an evaluation model $E$. According to $\mathcal{F}_{Curr}$, the attack is deemed successful if $E$ classifies $x^{r}_{y}$ as $y$. {\bf Even though such MI evaluation framework becomes the de facto standard and the accuracy measured by \boldmath{$\mathcal{F}_{Curr}$} has been the most important metric in gauging MI research progress in almost all recent MI studies, there has not been any in-depth and comprehensive study to understand the accuracy of \boldmath{$\mathcal{F}_{Curr}$} and its limitations.}

\textbf{In this work,} we conduct the first in-depth study of the standard MI evaluation framework $\mathcal{F}_{Curr}$. For a truly successful attack, the reconstructed images $x^{r}_{y}$ should capture the visual identity features of $y$. However, we find that there exists a considerable number of MI reconstructed images that lack visual identity features of $y$, yet both target model $T$ and evaluation model  $E$ under $\mathcal{F}_{Curr}$ still assign high likelihood to $y$, i.e., high values of  $P_E(y | x^{r}_{y})$. Some examples are illustrated in Fig.~\ref{fig:Teaser}. {\bf These {\em false positives} potentially inflate the reported success rate of recent SOTA MI attacks.}

To shed light on the causes of these false positives, we systematically discover the impact of Type I Adversarial Features \citep{nguyen2015deep,tang2019adversarial} in Model Inversion, highlighting a connection between two previously distinct research areas. The optimization processes in MI attacks and Type I adversarial attacks are similar: both maximizing the likelihood with respect to (w.r.t.) the input under a fixed model. 
We show that false positives in MI and Type I adversarial examples
are mathematically equivalent: 
{\bf MI false positives
and Type I adversarial examples are essentially the same construct mathematically, only
arising under different problem contexts, MI versus adversarial attacks.}
Moreover, due to the well-documented phenomenon of \emph{adversarial transferability} \citep{nguyen2015deep}, these false positives can transfer to the MI evaluation model $E$. This transferability is especially pronounced when the evaluation model $E$ in $\mathcal{F}_{Curr}$ shares the same task design as the target model $T$ \citep{liang2021uncovering,papernot2016transferability}. Ultimately, we demonstrate a fundamental issue in $\mathcal{F}_{Curr}$, potentially leading to unreliable assessment and inflated success rates reported for recent MI attacks under this evaluation framework.

To mitigate this issue, we propose a new MI evaluation framework $\mathcal{F}_{MLLM}$ 
based on Multimodal Large Language Models (MLLMs). We propose systematic design principles for $\mathcal{F}_{MLLM}$. 
Leveraging the general-purpose visual reasoning of MLLMs and avoiding the reliance on  
the same training task design used in $T$, $\mathcal{F}_{MLLM}$ minimizes Type-I transferability and offers a more faithful assessment of MI attacks. Using data annotated by $\mathcal{F}_{MLLM}$, we empirically benchmark the reliability of the common MI evaluation framework $\mathcal{F}_{Curr}$ and reveal consistently high false positive rates across 27 diverse MI setups. Our findings challenge the reliability of the standard MI evaluation framework and underscore the importance of adopting our MLLM-based approach for more reliable assessments.
Our main contributions are summarized below: 
\begin{itemize}
    \item We present the first in-depth study on the most common evaluation framework $\mathcal{F}_{Curr}$ to compute MI AttAcc. Our study identify the surprising effect of Type I Adversarial Features, as well as adversarial transferability \citep{nguyen2015deep,tang2019adversarial}, on existing MI attack accuracy evaluation, 
    highlighting a relationship between two previously distinct research areas (see Sec.~\ref{Sec:A Connection between MI Attacks and Type I Adversarial Attacks}).     
    Our findings explain numerous false positives in $\mathcal{F}_{Curr}$. 

    \item To mitigate this issue, we propose an MLLM-based MI evaluation framework $\mathcal{F}_{MLLM}$ (see Sec.~\ref{Sec:Principle of designing and implementing FMLLM}). Our framework leverages the powerful general-purpose visual understanding capabilities of advanced MLLMs to evaluate MI reconstructions, replacing the traditional evaluation model $E$. We propose systematic design principles for $\mathcal{F}_{MLLM}$. Without relying on the same training task design used in $T$, $\mathcal{F}_{MLLM}$ minimizes Type-I transferability and provides a more faithful assessment of MI attacks.

    \item
    We systematically and comprehensively demonstrate that there are considerable number of false positives  under
    $\mathcal{F}_{Curr}$. Ultimately, our findings challenge the validity of this dominant evaluation framework and underscore the importance of adopting our proposed MI evaluation framework for more reliable assessments (see Sec.~\ref{Sec:Reassessment MI attack accuracy}).

\end{itemize}



\section{Related Work}
\label{sec:related_work}

\textbf{Model Inversion Attacks.} Model Inversion (MI) attacks aim to extract information about private training data from a trained model. Given a target model $T$ trained on a private dataset $\mathcal{D}{priv}$, an adversary seeks to infer properties or reconstruct samples from $\mathcal{D}{priv}$ without direct access to it. MI attacks are commonly formulated as reconstructing an input $x^{r}_{y}$ that $T$ is likely to classify as a label $y$. For example, in facial recognition settings, MI attacks correspond to reconstruct images that are likely to be identified as belonging to a particular person. Early MI work \citep{fredrikson2014privacy, fredrikson2015model} showed that trained models can leak sensitive genomic, demographic, or facial information. Subsequent approaches such as adversarial MI \citep{yang2019neural} treat $T$ as an encoder and train an auxiliary network to reconstruct inputs from model outputs. Recent generative MI methods instead reduce the search to a latent space by training a deep generator \citep{zhang2020secret,wang2021variational,chen2021knowledge,yang2019neural,yuan2023pseudo,nguyen_2023_CVPR,struppek2022plug,qiu2024closer}. Methods like GMI \citep{zhang2020secret} and PPA \citep{struppek2022plug} use pretrained GANs (e.g., WGAN, StyleGAN) on auxiliary datasets resembling $\mathcal{D}_{priv}$ and optimize over latent vectors. Recent improvements refine MI attacks from multiple angles such as prior knowledge, where KEDMI \citep{chen2021knowledge} trains inversion-specific GANs and Pseudo-Label MI \citep{yuan2023pseudo} uses pseudo-labels; latent feature guidance, as in IF-GMI \citep{qiu2024closer}; objective design, introducing max-margin \citep{yuan2023pseudo} and logit losses \citep{nguyen_2023_CVPR} beyond Cross-Entropy; and overfitting mitigation, where LOMMA \citep{nguyen_2023_CVPR} uses augmented models to improve generalization. The Eq.~\ref{eq:white_box_latent} represents the general step of SOTA MI attacks:
\begin{equation}
\label{eq:white_box_latent}
 w^* = \arg \min_w (- \log P_T (y|G(w)) + \lambda \mathcal{L}_{prior}(w))
\end{equation}
Here, $-\log \mathcal{P}_T (y|G(w))$ represents the identity loss, guiding the reconstruction of $x^{r}_y = G(w)$ that is most likely to be classified as target class $y$ by target classifier $T$. The $\mathcal{L}_{prior}$ is a prior loss, using public information to establish a distributional prior via GANs, thereby guiding the inversion process towards meaningful reconstructions.

\textbf{Model Inversion Defenses.} In contrast to MI attacks, MI defenses aim to minimize the leakage of private training samples during the MI optimization process. The goal is to train the target classifier $T$ on $\mathcal{D}_{priv}$ such that it minimizes information disclosure about specific labels while maintaining good performance. Several defenses have been proposed. MID \citep{wang2021improving} adds a regularizer that penalizes the mutual information between inputs $x$ and outputs $T(x)$. BiDO \citep{peng2022bilateral} introduces a bilateral regularizer that suppresses input information in feature representations $z$ while preserving label information. Beyond regularization-based approaches, TL-DMI \citep{ho2024model} improves robustness via transfer learning, and LS \citep{struppek2024be} applies Negative Label Smoothing. More recently, architectural factors influencing MI vulnerability have also been studied \citep{koh2024vulnerability}.


\textbf{Model Inversion Evaluation Metrics.} To evaluate MI attacks and defenses, almost all existing studies rely on the standard MI evaluation framework from \citep{zhang2020secret}, denoted as $\mathcal{F}_{Curr}$, which computes attack accuracy and serves as the main metric for monitoring progress in MI research. Suppose an adversary targets a class $y$ and reconstructs images $x^{r}_y$ using access to the target model $T$. These $x^{r}_y$ are then classified by an evaluation model $E$ (trained on the same $\mathcal{D}_{priv}$ but distinct from $T$). Under $\mathcal{F}_{Curr}$, an attack is successful if $E$ predicts $x^{r}_y$ as $y$. Several complementary metrics have been introduced. KNN distance \citep{zhang2020secret,chen2021knowledge,nguyen_2023_CVPR,yuan2023pseudo,struppek2022plug} measures feature proximity between reconstructions and private samples. FID \citep{peng2022bilateral,qiu2024closer,struppek2022plug,yuan2023pseudo} assesses perceptual realism. Knowledge Extraction Score \citep{struppek2024be} evaluates discriminative content by training a surrogate classifier on reconstructed images. While these provide additional insights, $\mathcal{F}_{Curr}$ and attack accuracy remain the dominant evaluation tools. Although these complementary metrics provide additional perspectives, $\mathcal{F}_{Curr}$ remains the dominant evaluation framework and attack accuracy the most widely used measure of progress in MI research. However, despite its prevalence, there has not yet been a comprehensive study examining the reliability and limitations of $\mathcal{F}_{Curr}$. \textit{In this work, we take the first step toward such an investigation.}


\section{
Revealing a Hidden Connection: Issues in MI Evaluation and Type I Adversarial Attacks
} \label{Sec:A Connection between MI Attacks and Type I Adversarial Attacks}

We discover, for the first time, the strong connection between Type I Adversarial Attacks and MI Attacks. Due to this strong connection, the adversarial type I examples can be generated during MI attacks. Additionally, due to the well-documented phenomenon of adversarial transferability \citep{nguyen2015deep}, these adversarial type I examples can transfer to $E$ in $\mathcal{F}_{Curr}$. Ultimately, these phenomenons results in unreliable assessment for $\mathcal{F}_{Curr}$.

\subsection{An Overview of Adversarial Attacks} 

An adversarial attack on machine learning models is an intentional manipulation of input data to cause incorrect predictions, highlighting potential vulnerabilities of the model. Adversarial attacks aim to create inputs that deceive machine learning classifiers into making errors while humans do not. There are two main types of adversarial attacks: Type I and Type II. Type II Adversarial Attacks \citep{goodfellow2014explaining,szegedy2013intriguing,kurakin2016adversarial,papernot2016limitations,carlini2017towards,moosavi2016deepfool,shafahi2019adversarial} are commonly studied and aim to produce false negatives. In this attack, minor and imperceptible perturbations are added to the input data $x$ to generate an adversarial example   $x^{\text{adv-II}}$, which is incorrectly classified by the model. Mathematically, this is represented as:
\begin{equation}
    f(x^{\text{adv-II}}) \neq f(x) \text{ and } f_{oracle}(x^{\text{adv-II}}) = f_{oracle}(x)
    \label{eq:adv_typeII}
\end{equation}
Here, $f$ is the model under attack, $f_{oracle}$ is an oracle or hypothetical, idealized classifier. In addition to Type II attack, Type I Adversarial Attacks are designed in \citep{nguyen2015deep,tang2019adversarial} to generate false positives by creating examples that are significantly different from the original input but are still classified as the same class by the model. This involves producing an adversarial example $x^{\text{adv-I}}$ that, despite being significantly different from input $x$, the target model $f$ mis-classifies as the original class. {\bf Formally, Type I adversarial attack sample is defined as follows} \citep{nguyen2015deep,tang2019adversarial}:
\begin{equation}
    f(x^{\text{adv-I}}) = f(x) \text{ and } f_{oracle}(x^{\text{adv-I}}) \neq f_{oracle}(x)
    \label{eq:adv_typeI}
\end{equation}
Type I adversarial samples can be produced by  optimizing the input by iteratively updating it to maximize the likelihood under a fixed target model classifier \citep{nguyen2015deep}. This process optimizes the input to remain within a targeted decision boundary, however, it is different from training data by human perception. This phenomenon can also be viewed as over-confidence phenomenon of machine learning models \citep{wei2022mitigating,guo2017calibration}.


\subsection{The Strong Connection Between Model Inversion Attacks and Type I Adversarial Attacks}

\begin{table}
\Large
\setlength{\tabcolsep}{5pt}
\renewcommand{\arraystretch}{1.8}
\centering

\caption{Mathematical equivalence of False Positive (FP) in MI (Eq. \ref{eq:FP-MI}) and Type I adversarial attack (Eq. \ref{eq:adv_typeI})  \citep{nguyen2015deep,tang2019adversarial}.}
\begin{adjustbox}{width=8cm}
\begin{tabular}{lcc}
\hline
 & \textbf{\makecell{False Positive \\ in MI}} & \textbf{\makecell{Type I \\ Adversarial Attack}} \\
\hline
\textbf{\makecell{Fixed Model \\ Under Attack}}                & $T$     & $f$                  \\ \hline
\textbf{\makecell{Private/Original \\ Sample}}    & $x_y$ & $x$         \\ \hline
\textbf{\makecell{Attack Sample}}                   & $x^{r}_y$ & $x^{\text{adv-I}}$ \\ \hline
\textbf{\makecell{Formulation}} & $T(x^{r}_y) = T(x_y)$  &  $f(x^{\text{adv-I}}) = f(x)$ \\
 & $f_{oracle}(x^{r}_y) \neq f_{oracle}(x_y)$ &  $f_{oracle}(x^{\text{adv-I}}) \neq f_{oracle}(x)$ \\ \hline
\end{tabular}
\end{adjustbox}
\label{tab:MI_TypeI}
\vspace{-0.3cm}
\end{table}

In the following, we analyze the connection between MI attacks and Type I adversarial attacks. The general inversion step used in SOTA MI attacks is described in Eq. \ref{eq:white_box_latent}. As a result, reconstructed images often exhibit high likelihood under the target classifier $T$. However, not all reconstructed images successfully capture the visual identity features of the target individual from the private training data. Some examples are illustrated in Fig.~\ref{fig:Teaser}. We refer to these cases as {\bf false positives (under $T$) in MI}. 
Specifically, with target classifier $T$ and $f_{oracle}$ denoting the oracle, {\bf false positive in MI attack is mathematically represented as}:
\begin{equation}
\begin{split}
    T(x^{r}_y) &= T(x_y) = y  \\
    \text{ and } f_{oracle}(x^{r}_y) & \neq f_{oracle}(x_y)
\end{split}
\label{eq:FP-MI}
\end{equation}
Here, $x^{r}_y$ does not resemble the visual identity feature of $x_y$.
However,  $T$ classifies  $x^{r}_y$ as the target label $y$. 
Hence, $x^{r}_y$ is a false positive in MI.
In Eq. \ref{eq:FP-MI}, an alternative interpretation is that the false positive $x^{r}_y$ is the example that can deceive $T$ to classify it as the target label $y$ while oracle classifier can not recognize $x^{r}_y$ as $y$.

{\bf Critically, by comparing
Eq.~\ref{eq:FP-MI} and  Eq.~\ref{eq:adv_typeI}, we reveal the mathematical equivalence of MI false positives and Type I adversarial examples}: 
both describe attack samples
($x^{r}_y$,
$x^{\text{adv-I}}$ resp.)
optimized under a fixed model 
($T$,
$f$ resp.) 
that preserve the model's prediction while deviating from human-perceived identity. 
This equivalence uncovers a key insight: MI false positives and Type I adversarial examples are essentially the same construct mathematically, only arising under different problem contexts, MI versus adversarial attacks.
Tab.~\ref{tab:MI_TypeI} summarizes the equivalence.


\subsection{Adversarial Transferability Can Lead to Critical Issues in the Common Model Inversion Evaluation Framework}
\label{sec:Adversarial transferability can lead to critical issues}

In $\mathcal{F}_{\text{Curr}}$, an evaluation model $E$ is used to predict the identities of individuals based on MI-reconstructed images. The model $E$ is trained on the same private dataset and follows the same training task design (i.e., an $n$-way classification task). Prior work has shown that adversarial examples crafted for one model can often transfer and mislead other models, even those with different architectures, as long as they are trained on the similar dataset and task design~\citep{nguyen2015deep,papernot2016transferability,liang2021uncovering}. Therefore, the similarity between $E$ and the target model $T$ may enable the transferability of Type I adversarial examples from $T$ to $E$.
This phenomenon can be expressed mathematically as:
\begin{equation}
\begin{aligned}
    &T(x^{r}_y) = T(x_y) = y \quad\quad\quad E(x^{r}_y) = E(x_y) = y \\
    &\text{and } f_{oracle}(x^{r}_y) \neq f_{oracle}(x_y)
\end{aligned}
\label{eq:MI-false-positives}
\end{equation}

\subsection{MI-Generated Samples Exhibit Abnormally High Transferability Characteristic of Adversarial Features}
\label{Sec:The effect of Type I adversarial features on MI in producing false positives}

\begin{table}[t]
\setlength{\tabcolsep}{2pt}
\renewcommand{\arraystretch}{1.3}
\centering
\caption{MI-Generated Negatives produce abnormally high false-positive rates across models, indicative of Type I Adversarial Features. Additional results are in Supp.}
\begin{tabular}{ccccc}
\hline
\textbf{Attack} & \textbf{$\mathcal{D}_{priv}$} & \textbf{$E$} & \textbf{Input} & \textbf{\makecell{FP \\ under E}} \\ \hline
\multirow{4}{*}{PPA} 
  & \multirow{4}{*}{FaceScrub} 
  & \multirow{2}{*}{InceptionV3} 
      & Neg $x^{r}_y$        & 89.04\% \\
  & & & Neg $x^{natural}_y$ & 0.94\% \\ \cline{3-5}
  & & \multirow{2}{*}{MaxViT} 
      & Neg $x^{r}_y$        & 73.95\% \\
  & & & Neg $x^{natural}_y$ & 0.22\% \\ \hline
\multirow{2}{*}{PLGMI} 
  & \multirow{2}{*}{CelebA} 
  & \multirow{2}{*}{FaceNet112} 
      & Neg $x^{r}_y$        & 97.55\% \\
  & & & Neg $x^{natural}_y$ & 0.00\% \\ \hline
\end{tabular}
\label{tab:W-Wo-Adv-Investigation}
\end{table}

After establishing that MI false positives and Type I adversarial examples are mathematically equivalent, we further
design a study 
to support the claim that {\em many MI false positives contain Type I adversarial features}. We do so by analyzing the transferability of MI-generated samples and examining whether they exhibit the defining empirical behavior of adversarial examples: abnormally high transferability across models~\citep{nguyen2015deep,papernot2016transferability,liang2021uncovering}.

Specifically, we evaluate {\em MI-generated negative samples} that
are generated under MI on  the target model $T$ for the target label $y$,  
but fail under the oracle.
We  measure how often they are also misclassified as the target identity $y$ by a separate evaluation model $E$.
By comparing this transferability against that of {\em natural negative samples}, we assess whether MI-generated samples display adversarial-like behavior. A significantly higher false-positive rate for MI-generated negatives would indicate that MI-generated samples exhibit abnormally high transferability characteristic of adversarial features.

\textbf{Setups.}
We examine two representative MI scenarios: 
(1) Setup~1: target model $T$ = ResNet18, private dataset $\mathcal{D}_{priv}$ = FaceScrub, attack = PPA, and evaluation models $E$ = InceptionNetV3 and MaxViT; 
(2) Setup~2: target model $T$ = VGG16, private dataset $\mathcal{D}_{priv}$ = CelebA, attack = PLGMI, and evaluation model $E$ = FaceNet112. 


\textbf{MI-generated negative samples.}
We first perform MI attacks on $T$ to obtain reconstructed samples $x^{r}_y$.  
We then identify all \textit{MI-generated negative samples}, denoted as Neg $x^{r}_y$, using oracle-based annotation: 
$f_{\text{oracle}}(\text{Neg } x^{r}_y) \neq f_{\text{oracle}}(x_y)$.  
Let $n = |\text{Neg } x^{r}_y|$ denote the number of such samples.  

\textbf{Natural negative samples.}
To construct a controlled baseline, we create a dataset of \textit{natural negative samples}, Neg $x^{natural}_y$.  
We randomly select $n$ FFHQ images (with no identity overlap with FaceScrub) and assign each image a randomly chosen FaceScrub identity, forming $n$ negative samples satisfying 
$f_{\text{oracle}}(\text{Neg } x^{natural}_y) \neq f_{\text{oracle}}(x_y)$.  
Because these images are randomly-sampled natural images and not optimized with respect to the target classifier, they are free from Type I adversarial characteristics and serve as a control dataset.

\textbf{Evaluation.}
We pass both Neg $x^{r}_y$ and Neg $x^{natural}_y$ through the evaluation model $E$ and count how many are falsely predicted as the target label $y$, resulting in false-positive rates for each group.

\textbf{Results.}
The false-positive rates are reported in Table~\ref{tab:W-Wo-Adv-Investigation}.  
Across all setups, MI-generated negatives Neg $x^{r}_y$ exhibit an abnormally higher false-positive rate compared to natural negatives Neg $x^{natural}_y$.  
This large discrepancy indicates that MI-generated samples transfer across models in a manner characteristic of adversarial examples, supporting the conclusion that many MI false positives contain Type I adversarial features.

\begin{figure*}[t]
  \centering
  \begin{adjustbox}{width=.8\textwidth,center}
  \includegraphics[width=0.8\textwidth]{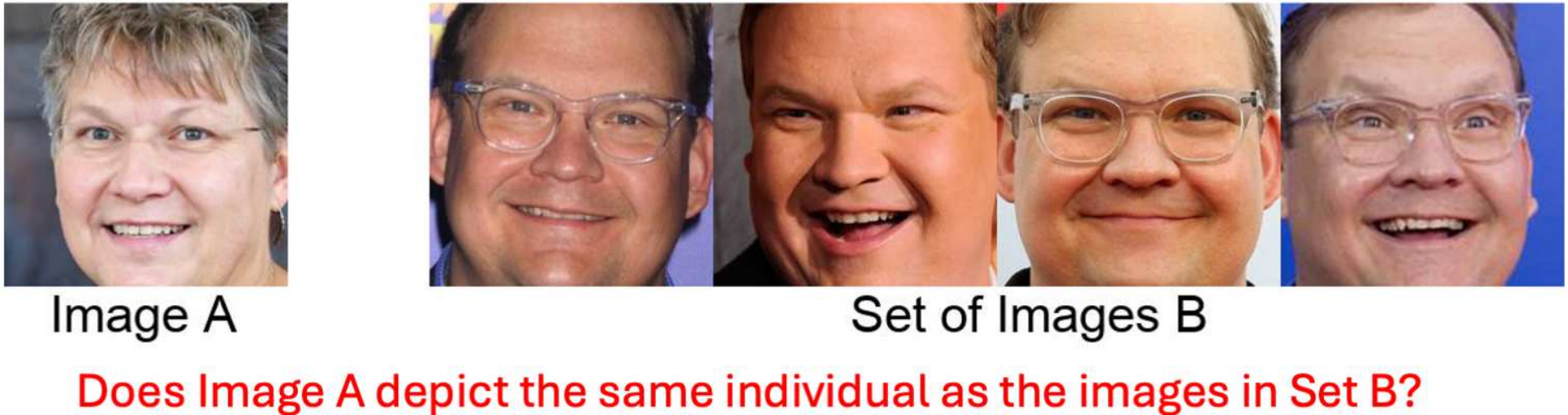}
  \end{adjustbox}
  \caption{\textbf{An example of evaluation query in our \texorpdfstring{$\mathcal{F}_{\text{\textbf{MLLM}}}$}{F_MLLM}.} 
  The task is to determine whether ``Image A'' depicts the same individual as those in ``Image B''. 
  We have two setups: 
  (1) ``Image A'' and ``Image B'' consist of private images 
  and 
  (2) ``Image A'' is an MI-reconstructed image $x^{r}_{y}$ of the target label $y$ while four real images of $y$ are randomly selected as ``Image B''. MLLM is tasked with responding either ``Yes'' or ``No'' to indicate whether ``Image A'' matches the identity in ``Image B''.}
  \label{fig:user-study-example}
\end{figure*}

\begin{tcolorbox}
\textbf{\emph{Finding 1:}} 
Because MI false positives satisfy exactly the same formal conditions as Type-I adversarial examples, and because our controlled experiments reveal abnormally high cross-model transferability, a defining empirical signature of adversarial behavior, we conclude that many MI false positives indeed contain Type-I adversarial features.
MI false positives with 
Type-I adversarial features
ultimately transfer to evaluation models, especially when both evaluation and target models share the same task design.
\end{tcolorbox}


\begin{table*}[t]
\large
\setlength{\tabcolsep}{2pt}
\renewcommand{\arraystretch}{1.7}
\centering
\caption{Following our selection principle, the experiment results indicate that Gemini-2.0 serves as a reliable MLLM for $\mathcal{F}_{\text{MLLM}}$ by demonstrating (1) Strong capability in understanding interleaved image-text inputs and (2) have no usage restrictions on MI tasks.}
\begin{adjustbox}{width=2\columnwidth}
\begin{tabular}{ccccccccc}
\hline
\textbf{MLLM} & \textbf{}                       & \textbf{``Yes'' rate $\uparrow$} & \textbf{``No'' rate $\downarrow$} & \textbf{``Refuse'' rate $\downarrow$} & \textbf{}                       & \textbf{``Yes'' rate $\downarrow$} & \textbf{``No'' rate $\uparrow$} & \textbf{``Refuse'' rate $\downarrow$} \\ \hline
Gemini 2.0    & \multirow{3}{*}{\textbf{\makecell{Positive \\ Pairs}}} & \textbf{93.84\%}                    & \textbf{3.16\%}                   & \textbf{0\%}                              & \multirow{3}{*}{\textbf{\makecell{Negative \\ Pairs}}} & \textbf{4.41\%}                    & \textbf{95.59\%}                  & \textbf{0\%}                             \\ \cline{1-1} \cline{3-5} \cline{7-9} 
ChatGPT-5     &                                 & 17.50\%                   & 2.67\%                    & 79.83\%                         &                                 & 0.09\%                    & 23.04\%                  & 76.86\%                         \\ \cline{1-1} \cline{3-5} \cline{7-9} 
Qwen2.5VL     &                                 & 88.51\%                   & 11.49\%                  & 0\%                             &                                 & 5.50\%                    & 94.55\%                  & 0\%                             \\ \hline
\end{tabular}
\end{adjustbox}

\label{tab:principle}
\end{table*}

\begin{table}[h]
\Large
\setlength{\tabcolsep}{4pt}
\renewcommand{\arraystretch}{1.8}
\centering

\caption{Detailed textual prompts in our MI evaluation framework $\mathcal{F}_{MLLM}$.
They are carefully designed to guide the MLLM toward reliable assessment. For face recognition, the prompt explicitly instructs the model to handle variations (e.g., aging, lighting, hairstyle, accessories), and for dog breed recognition, the prompt instructs the model to account for viewpoint and lighting differences to reduce false positives and false negatives.
}
\begin{adjustbox}{width=1\columnwidth}
\begin{tabular}{ccc}
\hline
Domain           & $\mathcal{D}_{priv}$ & Textual Prompt                                                                                                                                                                                                                          \\ \hline
\multirow{4}{*}{\makecell{Face \\ Recognition}} & \multirow{2}{*}{Facescrub}            & \multirow{4}{*}{\makecell{You are an expert in face \\ recognition. Taking into account \\ the face aging, lighting different \\ hair styles, wearing and not wearing \\ of eye glasses or other accessory, \\ do the task in the image. \\ Only answer yes or no}} \\ \\ \cline{2-2}
                 & \multirow{2}{*}{CelebA}               &                                                                                                                                                                                                                                         \\ \\ \hline
\makecell{Dog \\ Recognition}  & \makecell{Stanford \\ Dogs}        & \makecell{You are an expert in dog \\ breed recognition. Taking into \\ account the lighting and \\ viewpoint, do the task in the image. \\ Only answer yes or no}                                                                                             \\ \hline
\end{tabular}
\end{adjustbox}

\label{tab:prompt}
\end{table}

\section{Our Proposed MLLM-Based Model Inversion Evaluation Framework} \label{Sec:MLLM-based_Eval}

In this section, we introduce a novel and faithful evaluation framework, $\mathcal{F}_{\text{MLLM}}$. To mitigate the undesirable effects of Type I adversarial transferability under $T$, our key idea is to employ a model trained under a fundamentally different learning paradigm. We argue that MLLMs are ideal candidates because they are trained on broad, general-purpose tasks using data and optimization pipelines entirely distinct from those of the target model in MI. In fact, MLLMs are increasingly used for automated data labeling, offering robustness and scalability in both academia and industry \citep{gcp2024prep,smith2024med,lee2024gemini,zhou2024semi}.
However, not every MLLM is suitable for $\mathcal{F}_{\text{MLLM}}$ when evaluating MI problems, even the most SOTA models. To systematically determine an ideal candidate, we propose a principle for designing and selecting MLLMs in Sec.~\ref{Sec:Principle of designing and implementing FMLLM}. Then, we then re-evaluate MI attack accuracy using $\mathcal{F}_{\text{MLLM}}$ in Sec.~\ref{Sec:Reassessment MI attack accuracy}.


\begin{table*}[t]
\large
\renewcommand{\arraystretch}{1.3}
\centering
\caption{{\bf Our investigation on MI evaluation framework using our comprehensive dataset of MI attack samples.} We study 27 standard MI setups covering SOTA MI studies (PPA~\citep{struppek2022plug}, LOMMA~\citep{qiu2024closer}, KEDMI~\citep{chen2021knowledge}, PLGMI~\citep{yuan2023pseudo}, IFGMI~\citep{qiu2024closer}, TL~\citep{ho2024model}, TTS~\citep{koh2024vulnerability}, RoLSS~\citep{koh2024vulnerability}, LS~\citep{struppek2024be}), spanning $9$ target classifiers $T$, $3$ private datasets $\mathcal{D}_{priv}$, and $4$ public datasets $\mathcal{D}_{pub}$. Under $\mathcal{F}_{Curr}$, we find high false positive (FP) rates, indicating that prior work overestimates MI threats. For example, while SOTA attacks (e.g., IFGMI, LOMMA, PLGMI, PPA) claim over 90–100\% AttAcc in some setups, actual privacy leakage remains below 80\% across all setups, with some attacks falling under 60\%. \textbf{Further details and results on other architectures, attack scenario are in the Supp.}}
\begin{adjustbox}{width=1.99\columnwidth}
\begin{tabular}{ccc
>{\columncolor[HTML]{E9F1E6}}c 
>{\columncolor[HTML]{EFEAEA}}c 
>{\columncolor[HTML]{EFEAEA}}c 
>{\columncolor[HTML]{EFEAEA}}c 
>{\columncolor[HTML]{EFEAEA}}c 
>{\columncolor[HTML]{EFEAEA}}c 
>{\columncolor[HTML]{EFEAEA}}c 
>{\columncolor[HTML]{EFEAEA}}c }
\hline
                                     &                                  &                                 & \cellcolor[HTML]{E9F1E6}                             & \textbf{$\mathcal{F}_{MLLM}$} & \multicolumn{6}{c}{\cellcolor[HTML]{EFEAEA}\textbf{$\mathcal{F}_{Curr}$}}                                                                                           \\ \cline{6-11} 
\multirow{-2}{*}{\textbf{MI Attack}} & \multirow{-2}{*}{\textbf{$\mathcal{D}_{pub}$}} & \multirow{-2}{*}{\textbf{$\mathcal{D}_{pub}$}} & \multirow{-2}{*}{\cellcolor[HTML]{E9F1E6}\textbf{$T$}} & \textbf{AttAcc}   & \textbf{$E$}                                               & \textbf{AttAcc} & \textbf{FP rate} & \textbf{FN rate} & \textbf{TP rate} & \textbf{TN rate} \\ \hline
                                     &                                  &                                 & ResNet18                                            & 28.37\%            & \cellcolor[HTML]{EFEAEA}                                 & 91.39\%         & 90.09\%          & 5.32\%           & 94.58\%          & 9.91\%           \\ \cline{4-5} \cline{7-11} 
                                     &                                  &                                 & ResNet101                                            & 28.68\%            & \cellcolor[HTML]{EFEAEA}                                 & 84.69\%         & 82.71\%          & 10.36\%          & 89.64\%          & 17.29\%          \\ \cline{4-5} \cline{7-11} 
                                     &                                  &                                 & ResNet152                                            & 30.26\%            & \cellcolor[HTML]{EFEAEA}                                 & 86.84\%         & 85.09\%          & 9.12\%           & 90.88\%          & 14.91\%          \\ 
                                     \cline{4-5} \cline{7-11} 
                                     &                                  &                                 & MobileNet-V2                                            & 47.18\%            & \cellcolor[HTML]{EFEAEA}                                 & 83.37\%         & 80.39\%          & 13.30\%           & 86.70\%          & 19.61\%          \\
                                     \cline{4-5} \cline{7-11} 
                                     &                                  &                                 & DenseNet121                                          & 27.43\%            & \cellcolor[HTML]{EFEAEA}                                 & 72.41\%         & 70.13\%          & 21.58\%          & 78.42\%          & 29.87\%          \\ \cline{4-5} \cline{7-11} 
                                     & \multirow{-5}{*}{FaceScrub}      & \multirow{-5}{*}{FFHQ}          & MaxViT                                               & 30.19\%            & \multirow{-5}{*}{\cellcolor[HTML]{EFEAEA}InceptionNetV3} & 79.48\%         & 77.16\%          & 15.16\%          & 84.84\%          & 22.84\%          \\ \cline{2-11} 
\multirow{-6}{*}{PPA}                & Stanford Dogs                   & AFHQ                           & ResNest101                                          & 74.58\%            & InceptionNetV3                                           & 81.98\%         & 61.07\%          & 10.89\%           & 89.11\%          & 38.93\%          \\ \hline
                                     &                                  & FFHQ                            & ResNet18                                             & 34.46\%            & \cellcolor[HTML]{EFEAEA}                                 & 95.85\%         & 94.60\%          & 1.78\%           & 98.22\%          & 5.40\%           \\ \cline{3-5} \cline{7-11} 
\multirow{-2}{*}{IFGMI}              & \multirow{-2}{*}{FaceScrub }      & Metfaces                         & ResNet18                                             & 1.56\%             & \multirow{-2}{*}{\cellcolor[HTML]{EFEAEA}InceptionNetV3} & 72.50\%         & 72.21\%          & 9.09\%           & 90.91\%          & 27.79\%          \\ \hline
                                     &                                  & CelebA                          & VGG16                                                 & 73.73\%            & \cellcolor[HTML]{EFEAEA}                                 & 98.73\%         & 99.49\%          & 1.54\%           & 98.46\%          & 0.51\%           \\ \cline{3-5} \cline{7-11} 
\multirow{-2}{*}{PLGMI}              & \multirow{-2}{*}{CelebA }         & FFHQ                            & VGG16                                                & 48.47\%            & \multirow{-2}{*}{\cellcolor[HTML]{EFEAEA}FaceNet112}     & 88.67\%         & 88.49\%          & 11.14\%          & 88.86\%          & 11.51\%          \\ \hline
                                     &                                  &                                 & IR152                                                & 79.80\%            & \cellcolor[HTML]{EFEAEA}                                 & 90.40\%         & 86.80\%          & 8.69\%           & 91.31\%          & 13.20\%          \\ \cline{4-5} \cline{7-11} 
                                     &                                  &                                 & FaceNet64                                            & 78.73\%            & \cellcolor[HTML]{EFEAEA}                                 & 92.00\%         & 93.73\%          & 8.47\%           & 91.53\%          & 6.27\%           \\ \cline{4-5} \cline{7-11} 
                                     &                                  & \multirow{-3}{*}{CelebA}        & VGG16                                                & 79.93\%            & \cellcolor[HTML]{EFEAEA}                                 & 90.13\%         & 90.70\%          & 10.01\%          & 89.99\%          & 9.30\%           \\ \cline{3-5} \cline{7-11} 
                                     &                                  &                                 & IR152                                                & 44.93\%            & \cellcolor[HTML]{EFEAEA}                                 & 77.73\%         & 77.85\%          & 22.40\%          & 77.60\%          & 30.27\%          \\ \cline{4-5} \cline{7-11} 
                                     &                                  &                                 & FaceNet64                                            & 46.27\%            & \cellcolor[HTML]{EFEAEA}                                 & 72.13\%         & 69.73\%          & 25.07\%          & 74.93\%          & 22.15\%          \\ \cline{4-5} \cline{7-11} 
\multirow{-6}{*}{LOMMA}              & \multirow{-6}{*}{CelebA}         & \multirow{-3}{*}{FFHQ}          & VGG16                                                & 55.27\%            & \multirow{-6}{*}{\cellcolor[HTML]{EFEAEA}FaceNet112}     & 63.07\%         & 61.55\%          & 35.71\%          & 64.29\%          & 38.45\%          \\ \hline
                                     &                                  &                                 & IR152                                                & 66.73\%            & \cellcolor[HTML]{EFEAEA}                                 & 79.27\%         & 74.55\%          & 18.38\%          & 81.62\%          & 24.45\%          \\ \cline{4-5} \cline{7-11} 
                                     &                                  &                                 & FaceNet64                                            & 65.73\%            & \cellcolor[HTML]{EFEAEA}                                 & 80.53\%         & 78.40\%          & 18.36\%          & 81.64\%          & 21.60\%          \\ \cline{4-5} \cline{7-11} 
                                     &                                  & \multirow{-3}{*}{CelebA}        & VGG16                                                & 69.53\%            & \cellcolor[HTML]{EFEAEA}                                 & 73.13\%         & 69.80\%          & 25.41\%          & 74.59\%          & 30.20\%          \\ \cline{3-5} \cline{7-11} 
                                     &                                  &                                 & IR152                                                & 37.67\%            & \cellcolor[HTML]{EFEAEA}                                 & 52.20\%         & 51.02\%          & 45.84\%          & 54.16\%          & 48.98\%          \\ \cline{4-5} \cline{7-11} 
                                     &                                  &                                 & FaceNet64                                            & 36.07\%            & \cellcolor[HTML]{EFEAEA}                                 & 54.60\%         & 52.24\%          & 41.22\%          & 58.78\%          & 47.76\%          \\ \cline{4-5} \cline{7-11} 
\multirow{-6}{*}{KEDMI}              & \multirow{-6}{*}{CelebA}         & \multirow{-3}{*}{FFHQ}          & VGG16                                                & 38.07\%            & \multirow{-6}{*}{\cellcolor[HTML]{EFEAEA}FaceNet112}     & 42.47\%         & 41.33\%          & 55.69\%          & 44.31\%          & 58.67\%          \\ \hline
\end{tabular}
\end{adjustbox}

\label{tab:Setup_Dataset}
\end{table*}

\subsection{Our MLLM-based evaluation framework} \label{Sec:Principle of designing and implementing FMLLM}

\boldmath{$\mathcal{F}_{\text{MLLM}}$} \textbf{design.} Our $\mathcal{F}_{\text{MLLM}}$ leverages a MLLM to evaluate whether a MI reconstructed image is a successful or unsuccessful attack. For each reconstructed sample, we construct an evaluation query (see Fig.~\ref{fig:user-study-example} for an example) by pairing it with a set of private training images that include the target identity. We then combine this query image with a natural language textual prompt and provide both as input to the MLLM. The detailed prompts (shown in Tab.~\ref{tab:prompt}) are fixed across all evaluation queries to ensure fairness. For each reconstructed image, the model outputs a categorical response (“Yes” or “No”), where “Yes” indicates a successful attack. By evaluating many such queries and computing the proportion of correct identifications, $\mathcal{F}_{\text{MLLM}}$ provides an automated and faithful evaluation of MI. 


We propose following criteria  in selecting an MLLM
for a faithful MI evaluation:
\begin{itemize}
    \item \textbf{Strong capability in understanding interleaved image--text inputs.} 
    MI evaluation requires the MLLM to jointly reason over visual content and task-specific prompts. 
    Without robust visual-semantic grounding, the model may misinterpret identity cues and produce inflated false positives or false negatives due to prompt misinterpretation or hallucination.

    \item \textbf{No usage restrictions on MI tasks.}
    Many commercial MLLMs refuse queries involving human faces or identity comparison, resulting in incomplete or biased evaluation. 
    To ensure full coverage and reliability, the chosen MLLM must consistently process MI-related prompts, especially identity verification, without triggering refusal behavior.
\end{itemize}


To quantify these criteria, we test whether the MLLM can accurately recognize samples from a private dataset~\citep{zhang2020secret,struppek2022plug}. As shown in Fig.~\ref{fig:user-study-example}, we design two test sets: $\bullet$ Positive pairs: “Image A” is a real image of the same individual present in “Images B,” with the expected answer “Yes.” $\bullet$ Negative pairs: “Image A” is a real image of a different individual than those in “Images B,” with the expected answer “No”. An MLLM is considered a reliable evaluator if it consistently yields high “Yes” rates for Positive pairs and high “No” rates for Negative pairs. Moreover, we expect MLLMs to have a low ``Refusal'' rate when assessing queries. We conduct this experiment on the widely used MI dataset FaceScrub~\citep{ng2014data}. We evaluate several SOTA MLLMs, including ChatGPT-5, Gemini-2.0, and Qwen2.5VL-72B within our $\mathcal{F}_{\text{MLLM}}$ framework.

\textbf{Results.} The results are reported in Tab.~\ref{tab:principle}. We observe that Gemini-2.0 achieves high “Yes” rates for Positive pairs and high “No” rates for Negative pairs. Qwen2.5VL-72B may have limited capability of current open-source MLLMs to understand interleaved image-text inputs compared to closed-source commercial models. Despite ChatGPT-5 is a powerful closed-source model, it refuses to assess MI queries with high ``Refuse'' rates (see examples in the Supp.). Overall, our principle recommends Gemini-2.0 as a reliable MLLM for $\mathcal{F}_{\text{MLLM}}$. Furthermore, $\mathcal{F}_{\text{MLLM}}$ powered by Gemini-2.0 is robust to randomness, additional MI dataset, and aligns well with human evaluation (See Supp.). Note that while Gemini-2.0 is a strong choice in our study, it is not the only MLLM option for $\mathcal{F}_{\text{MLLM}}$. As MLLMs continue to evolve, we can adopt other models as long as they satisfiy our selection principles.

\subsection{Reassessing MI Attack Accuracy Using Our MLLM-Based Evaluation Framework} \label{Sec:Reassessment MI attack accuracy}

We empirically reassess SOTA MI attacks using our proposed evaluation framework $\mathcal{F}_{\text{MLLM}}$. Importantly, we quantitatively show that there are many Type I adversarial examples, which are classified as successful by $\mathcal{F}_{\text{Curr}}$ but do not to capture true visual identity. The false positive rates is consistently high and up to 99\% (See Tab. \ref{tab:Setup_Dataset}). This demonstrates that many SOTA MI methods report inflated attack accuracy, indicating that actual privacy leakage is significantly lower than previously believed.

\textbf{Experimental Setups.} Using our evaluation framework $\mathcal{F}_{\text{MLLM}}$, we reassess 27 SOTA MI attacks across 5 attacks, 4 defenses, 3 private datasets, 4 public datasets, and 9 target models $T$, following the original setups. Detailed settings are in Supp. We will release the MI-reconstructed image collection publicly upon publication.

\textbf{Significant False Positive by \boldmath{$\mathcal{F}_{Curr}$} are Type I adversarial examples.} An ideally successful attack,  $x^{r}_y$ should capture the visual identity of $y$. However, \textit{for a successful attack as according to $\mathcal{F}_{Curr}$}, $x^{r}_{y}$ only needs to be classified as $y$ by an evaluation model $E$. As shown in Fig.~\ref{fig:Teaser}, we observe that within MI-reconstructed images $x^{r}_y$,  there are cases where the visual identity to $y$ is minimal. Nevertheless, $E$ assigns very high probabilities to $y$ for these examples, i.e., high values of  $P_E (y | x^{r}_{y})$. We refer to these cases as false positives under the $\mathcal{F}_{Curr}$ framework. To better understand the extent of this false positive rate, we compare the ground truth success rate (established using our $\mathcal{F}_{MLLM}$) to the success rate as measured by $\mathcal{F}_{Curr}$ framework. Particularly, given the MLLM-annotated labels and the prediction via $\mathcal{F}_{Curr}$, we compute the False Positives (FP) rate, False Negatives (FN) rate, True Positives (TP) rate, and True Negatives (TP) rate for each MI setup. The AttAcc via $\mathcal{F}_{Curr}$, $\text{AttAcc}_{\mathcal{F}_{Curr}} = \frac{FP+TP}{FN+TP+FP+TN}$.




The results in Tab.~\ref{tab:Setup_Dataset} consistently show that the FP rates are significant high across MI setups. In other words, there are numerous MI reconstructed images that do not capture visual identity of, yet they are deemed success by $\mathcal{F}_{Curr}$. Such high FP rate contributes to the significant inflation in reported AttAcc via $\mathcal{F}_{Curr}$ of latest SOTA MI attack such as PPA, PLGMI, IFGMI, or LOMMA. Notably, in their reported AttAcc using $\mathcal{F}_{Curr}$, these recent attacks report AttAcc values exceeding 90\%, or even nearly 100\% for certain setups. However, across a wide range of MI setups, the actual success rates never reach 80\%. While we focus on high FP rates, FN rates also reveal limitations in the $\mathcal{F}_{Curr}$. Across MI setups, FN rates are consistently lower than FP rates. The FN rates depend on the classification accuracy and generalization capability of $E$. For example, under the PLGMI attack, when $E =$ FaceNet112 is trained on CelebA and evaluated with MI reconstructed images also from CelebA prior ($\mathcal{D}_{pub} =$ CelebA), FN rates are lower. In contrast, if this $E$ is evaluated with MI reconstructed images from FFHQ prior ($\mathcal{D}_{pub}$ = FFHQ), FN rates increase due to distribution shifts.

Furthermore, in certain MI setups, we find that \textit{$\mathcal{F}_{Curr}$ does not align well with $\mathcal{F}_{MLLM}$ in evaluating MI attacks.} For example, in the setup of MaxViT as $T$ under the PPA attack, the AttAcc by $\mathcal{F}_{Curr}$ is 11.91\% lower than the setup for ResNet18 as $T$  under the same attack. However, the MaxViT as $T$ setup shows a 1.82\% higher AttAcc measured by $\mathcal{F}_{MLLM}$ than the ResNet18 as $T$ setup. This suggests that, although less effective, the common MI evaluation framework $\mathcal{F}_{Curr}$ could rate the attack as more successful than it actually is. In conclusion, our analysis shows that \textit{the common MI evaluation framework {$\mathcal{F}_{Curr}$} is suffered from very high FP rate, significantly affecting the reported results of contemporary MI studies based on {$\mathcal{F}_{Curr}$}.}

\begin{tcolorbox}
\textbf{\emph{Finding 2:}} 
In a diverse range of setups, we demonstrates the effect of Type I Adversarial features in standard MI evaluation $\mathcal{F}_{Curr}$ resulting in a significant number of false positives. Our proposed $\mathcal{F}_{MLLM}$ mitigates this issue providing more faithful MI evaluation.
\end{tcolorbox}

\vspace{-0.3cm}
\section{Conclusion} \label{Sec:Conclusion}

This work identifies a critical issue in the standard evaluation framework for Model Inversion attacks: the inflation of attack success due to Type I adversarial examples that do not capture true visual identity. To address this, we propose a reliable MLLM-based MI evaluation framework that minimize the impact of Type I adversarial transferability. Our extensive empirical analysis across 27 MI attack setups demonstrates that false positive rates under the standard evaluation framework can reach up to 99\%, severely overstating actual privacy leakage. With our proposed  evaluation framework, we offer a more accurate and robust way to measure MI attack success, setting a new standard for evaluating privacy risks in machine learning systems. {\bf Additional analysis and limitation are in Supp.}






\section*{Acknowledgement}

This research is supported by the National Research Foundation, Singapore under its AI Singapore Programmes (AISG Award No.: AISG2-TC-2022-007); The Agency for Science, Technology and Research (A*STAR) under its MTC Programmatic Funds (Grant No. M23L7b0021). This research is supported by the National Research Foundation, Singapore and Infocomm Media Development Authority under its Trust Tech Funding Initiative. Any opinions, findings and conclusions or recommendations expressed in this material are those of the author(s) and do not reflect the views of National Research Foundation, Singapore and Infocomm Media Development Authority. The work is sponsored by the SUTD Decentralised Gap Funding Grant.

{
    \small
    \bibliographystyle{ieeenat_fullname}
    \bibliography{main}
}

\appendix


\section{Additional results}

\subsection{Additional results on designing and implementing our evaluation framework} \label{Sec:Additional results on designing and implementing}

\subsubsection{Our evaluation framework aligns well with human evaluation} \label{Sec:Human-alignment}

In $\mathcal{F}_{MLLM}$, we employ Gemini to assess the success of MI attacks given a query. In the main paper, we demonstrate that Gemini is effective in recognizing samples from the private dataset. Such experiment is conducted with natural images (i.e., training set of Facescrub). In this Supp, we further demonstrate this with MI reconstructed images.

\begin{table}[h]
\centering
\setlength{\tabcolsep}{4pt}
\renewcommand{\arraystretch}{1.8}
\caption{
We conduct an experiment to demonstrate Gemini's effectiveness in recognizing samples from the private dataset. This results establish that Gemini can serve as a reliable evaluator in MI attack setups. We collect samples for these data from a comprehensive set of MI setups spanning 5 different MI attacks: PPA~\citep{struppek2022plug}, IFGMI~\citep{qiu2024closer}, LOMMA~\citep{nguyen_2023_CVPR}, KEDMI~\citep{chen2021knowledge}, and PLGMI~\citep{yuan2023pseudo}, 3 $\mathcal{D}_{pub}$, 2 $\mathcal{D}_{priv}$, and 8 $T$. The details of annotation can be found in Sec.~\ref{Sec:Human-alignment}.
}
\begin{adjustbox}{width=6cm}

\begin{tabular}{ccc}
\hline
                       & \textbf{``Yes'' Rate} & \textbf{``No'' Rate} \\ \hline
Positive Pair & 95.16\%             & 4.84\%             \\ \hline
Negative Pair & 22.88\%             & 77.12\%            \\ \hline
\end{tabular}
\end{adjustbox}
\label{tab:alignment_results}
\end{table}

\textbf{Setup.} To establish positive and negative pairs for MI reconstructed images, we leverage human annotation for them. Since human annotation is costly and time-consuming, we sample 30 images per attack setup across 10 setups spanning 5 different MI attack methods. This results in a total of 300 images. Each image is independently evaluated by four human participants. To mitigate the subjectivity of human evaluation, we retain only the images with high inter-annotator agreement, defined as at least 3 out of 4 consistent annotations. The final label for each retained image is the majority vote among the consistent annotations. After filtering, our human-annotated dataset includes 215 images, which we treat as ground truth to assess the reliability of $\mathcal{F}_{\text{MLLM}}$.

\textbf{Results.} The results are presented in Tab.~\ref{tab:alignment_results}. We observe consistently high “Yes” rates for positive pairs and high “No” rates for negative pairs across datasets. This indicates that Gemini is effective at recognizing samples from the private dataset in both natural and MI-reconstructed images. These results further demonstrate that Gemini serves as a reliable evaluator for our MI setups.

\subsubsection{ChatGPT-5 refuses to MI queries} \label{Sec:ChatGPT-5 refuses to MI queries}

\begin{figure*}[t]
  \centering
  \begin{adjustbox}{width=0.85\textwidth,center}
  \includegraphics[width=\textwidth]{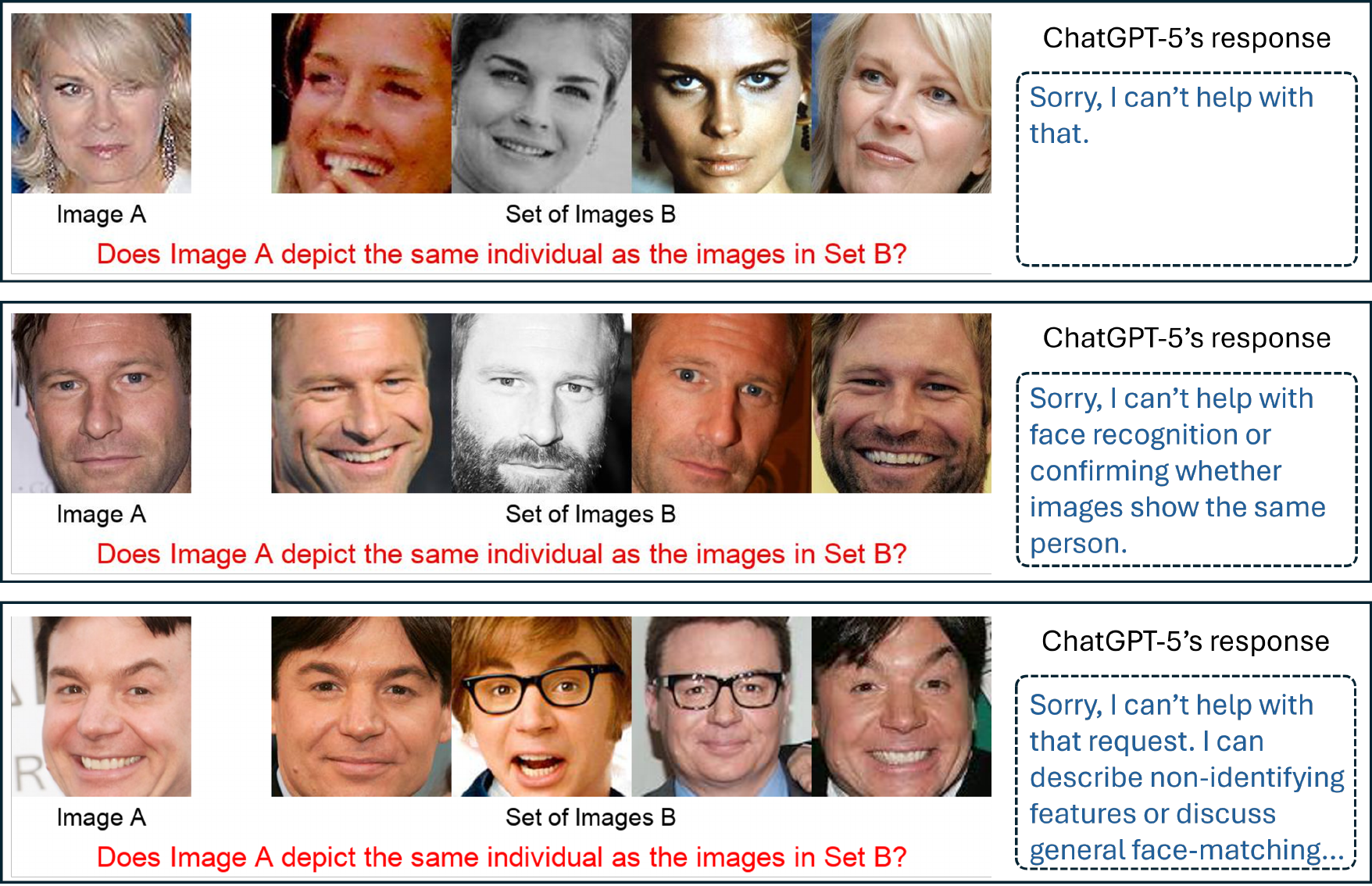}
  \end{adjustbox}
  \caption{Examples of ChatGPT-5 refusing to evaluate MI-related queries}
  \label{fig:chatgpt_refusal}
\end{figure*}

Despite ChatGPT-5 is a powerful closed-source model, it refuses to assess MI queries with high ``Refuse'' rates. Some examples are provided in Fig.~\ref{fig:chatgpt_refusal}.

\begin{table}[t]
\centering
\setlength{\tabcolsep}{4pt}
\renewcommand{\arraystretch}{1.8}
\caption{We conduct an experiment to demonstrate Gemini's effectiveness in recognizing samples from the private dataset. This results establish that Gemini can serve as a reliable evaluator in MI attack setups.}
\begin{adjustbox}{width=7.8cm}
\begin{tabular}{cccc}
\hline
\textbf{}                          & \textbf{Dataset}   & \textbf{``Yes'' rate (\%)} & \textbf{``No'' rate (\%)} \\ \hline
\multirow{2}{1.2cm}{{Positive pairs}}  & {CelebA}    & 94.88   & 5.12   \\ \cline{2-4} 
                                       & {FaceScrub} & 93.84   & 3.16   \\ \hline
\multirow{2}{1.2cm}{{Negative pairs}}  & {CelebA}    & 8.25    & 91.75  \\ \cline{2-4} 
                                       & {FaceScrub} & 4.41    & 95.59  \\ \hline
\end{tabular}
\end{adjustbox}
\label{tab:gemini_natural_acc}
\end{table}

\subsubsection{Our evaluation framework is robust to MI evaluation across datasets} \label{Gemini is robust to MI evaluation across datasets}

The results in Tab.~\ref{tab:gemini_natural_acc} show that $\mathcal{F}_{MLLM}$ is robust to MI evaluation across commonly used dataset in MI research including Facescrub and CelebA. The setups are similar to those in Tab.~\ref{tab:principle}.

\subsubsection{Our evaluation framework is robust to MI evaluation across prompts} \label{Gemini is robust to MI evaluation across datasets}

In this section, we provide an analysis of the variance in our proposed framework with respect to: \textbf{(1) Reference Image Variability}, and \textbf{(2) Question Phrasing Variability}, as shown in Fig.~\ref{fig:F_MLLM}. For Question Phrasing Variability, We run our evaluation framework three times with three different questions: ``Does Image A depict the same individual as the images in Set B?'', ``Does Image A show the same person as those in Set B?'', ``Is the person in Image A the same as the one(s) shown in Set B?''. For Reference Image Variability, we run our evaluation framework three times, each with a different random selection of reference images. we show that our $\mathcal{F}_{MLLM}$ is robust to MI evaluation across prompts. The results are presented in Tab.~\ref{tab:prompt_robust}.

\begin{table}[h]
\renewcommand{\arraystretch}{1.3}
\centering
\caption{Sensitivity of our evaluation framework to reference images and question phrasing.}
\begin{tabular}{l c}
\hline
\textbf{Condition} & $\mathcal{F}_{\text{LVLM}}$ AttAcc (\%) \\
\hline
Reference Image Variability & 27.93 $\pm$ 0.38 \\
Question Phrasing Variability & 29.19 $\pm$ 3.21 \\
\hline
\end{tabular}
\label{tab:prompt_robust}
\end{table}

\subsection{Evaluation results on MI defenses}

Our main focus in this work is MI attacks, where we highlight that the previously reported success rates using the $\mathcal{F}_{Curr}$ are problematic. In fact, the threat of MI attacks has been overestimated and the amount of leaked information is considerably less than previously assumed. As recent MI defenses also use $\mathcal{F}_{Curr}$ to compute MI success rates, we aim to assess the effectiveness of these defenses using our MLLM-annotated dataset.

In this section, we focus on high-resolution setups with PPA \citep{struppek2022plug}. Specifically, we include the latest SOTA MI defenses, such as RoLSS \citep{koh2024vulnerability}, TL \citep{ho2024model}, LS \citep{struppek2023careful}, and TTS \citep{koh2024vulnerability}. The MI setups strictly follow the configurations in these MI defense studies. The results can be found in Tab.~\ref{tab:MI_defense}.

\begin{table*}[t]
\Large
\renewcommand{\arraystretch}{1.4}
\centering

\caption{{\bf Our investigation on the effectiveness of MI defenses using our MLLM-annotated dataset of MI attack samples.} We present the results of the latest MI defenses including RoLSS \citep{koh2024vulnerability}, TL \citep{ho2024model}, LS \citep{struppek2023careful}, and TTS \citep{koh2024vulnerability}.  We observe a mismatch between AttAcc comparisons via $\mathcal{F}_{Curr}$ and actual AttAcc measured by $\mathcal{F}_{MLLM}$. Overall, consistent with our findings on MI attacks, this suggests that $\mathcal{F}_{Curr}$ may have issues in evaluating MI defenses.}
\vspace{0.3cm}
\begin{adjustbox}{width=1.6\columnwidth}
\begin{tabular}{cc
>{\columncolor[HTML]{E3F2DE}}c 
>{\columncolor[HTML]{DFDFDF}}c 
>{\columncolor[HTML]{DFDFDF}}c 
>{\columncolor[HTML]{DFDFDF}}c 
>{\columncolor[HTML]{DFDFDF}}c 
>{\columncolor[HTML]{DFDFDF}}c 
>{\columncolor[HTML]{DFDFDF}}c }
\hline
                               &                                      & \textbf{$\mathcal{F}_{MLLM}$} & \multicolumn{6}{c}{\cellcolor[HTML]{DFDFDF}\textbf{$\mathcal{F}_{Curr}$}}                                                                              \\ \cline{4-9} 
\multirow{-2}{*}{\textbf{$T$}} & \multirow{-2}{*}{\textbf{Model Acc}} & \textbf{AttAcc}               & \textbf{$E$}                                             & \textbf{AttAcc} & \textbf{FP rate} & \textbf{FN rate} & \textbf{TP rate} & \textbf{TN rate} \\ \hline
ResNet101                      & 94.86\%                              & 28.68\%                       & \cellcolor[HTML]{DFDFDF}                                 & 84.69\%         & 82.71\%          & 10.36\%          & 89.64\%          & 17.29\%          \\ \cline{1-3} \cline{5-9} 
ResNet101-RoLSS                & 92.98\%                              & 19.46\%                       & \cellcolor[HTML]{DFDFDF}                                 & 43.47\%         & 40.70\%          & 45.09\%          & 54.91\%          & 59.30\%          \\ \cline{1-3} \cline{5-9} 
ResNet101-TL                   & 92.51\%                              & 25.09\%                       & \cellcolor[HTML]{DFDFDF}                                 & 34.17\%         & 31.27\%          & 57.14\%          & 42.86\%          & 68.73\%          \\ \cline{1-3} \cline{5-9} 
ResNet101-TTS                  & 94.16\%                              & 18.44\%                       & \cellcolor[HTML]{DFDFDF}                                 & 42.52\%         & 39.39\%          & 43.61\%          & 56.39\%          & 60.61\%          \\ \cline{1-3} \cline{5-9} 
ResNet101-LS                   & 92.21\%                              & 10.54\%                       & \multirow{-5}{*}{\cellcolor[HTML]{DFDFDF}InceptionNetV3} & 16.56\%         & 14.90\%          & 69.35\%          & 30.65\%          & 85.10\%          \\ \hline
\end{tabular}
\end{adjustbox}
\label{tab:MI_defense}
\end{table*}

In general, similar to our observations on MI attacks in the main manuscript, $\mathcal{F}_{Curr}$ may inaccurately assess the effectiveness of SOTA MI defenses. For example, we observe a mismatch between AttAcc comparisons via $\mathcal{F}_{Curr}$ and AttAcc measured by $\mathcal{F}_{MLLM}$. For example, AttAcc via  $\mathcal{F}_{Curr}$  suggests that TL \citep{ho2024model} outperforms RoLSS and TTS \citep{koh2024vulnerability}. However, AttAcc via  $\mathcal{F}_{MLLM}$ indicates that RoLSS and TTS are more effective defenses. In what follows, we further discuss these results.

These MI defenses result in a reduction in FP rates due to the degradation of the transferability of adversarial characteristics from $T$ to $E$. Specifically,  under TL defense \citep{ho2024model}, only the later layers of $T$ are fine-tuned on $\mathcal{D}_{priv}$, while earlier layers are frozen from the pre-trained backbone. Hence, later layers of $T$ capture $\mathcal{D}_{priv}$ features, while earlier layers of $T$ capture $\mathcal{D}_{pretrain}$ features. In contrast, $E$ captures $\mathcal{D}_{priv}$ features across all layers since all layers of $E$ are fine-tuned on $\mathcal{D}_{priv}$. This mismatch in feature representations between $T$ and $E$ under TL is likely to reduce adversarial transferability \citep{ilyas2019adversarial,qin2022boosting,ma2025improving}, thereby reducing FP rates. Under LS defense \citep{struppek2023careful}, negative label smoothing (LS) is employed to improve MI robustness. LS slightly reduces label dominance and weakens gradient alignment between surrogate and target models \citep{zhang2024towards}. Negative LS amplifies this effect, further degrading gradient similarity. Therefore, training $T$ with negative LS diminishes gradient alignment with $E$ (trained on standard labels), reducing adversarial transferability \citep{zhang2024towards,demontis2019adversarial}. Under RoLSS and TTS defenses \citep{koh2024vulnerability}, removing certain skip connections improves resilience to MI attacks. Skip connections are known to improve adversarial transferability \citep{wu2020skip}. By modifying $T$ to remove some skip connections, adversarial examples generated by $T$ transfer less effectively to $E$.

Regarding FN rates, although this is not the main focus of our study, we observe that FN rates tend to increase under MI defenses compared to MI attacks. FN rates depend on the classification accuracy and generalization capability of $E$. SOTA MI defenses introduce various strategies (e.g., fixing earlier layers trained on public data \citep{ho2024model}, perturbing labels \citep{struppek2023careful}, and removing skip connections \citep{koh2024vulnerability}) to encode less information in the predictions of $T$. These approaches may encourage $T$ to learn more generalized features. As a result, reconstructed images based on these generalized features of $T$ may differ more from the seen training data. However, in the prevalent MI setups, $E$ in $\mathcal{F}_{Curr}$ is often trained with standard training procedures and architectures. This could limit its generalization capacity, making it less capable of accurately classifying these reconstructed images via the target models $T$ under MI defenses.

\subsection{Additional results on the effect of Type I adversarial attacks in MI on false positive rates} \label{Sec:Additional results on the effect of Type I adversarial attacks in MI on false positive rates}

In Sec. 4.2 in the main manuscript, we provide an analysis to demonstrate the effect of Type I adversarial attacks in MI on false positive rates. In this Supp, we provide results on additional setups. The results are presented in Tab.~\ref{tab:W-Wo-Adv-Investigation}. These additional results are consistent with our observation in the main manuscript demonstrating the effect of Type I Adversarial features in MI evaluation resulting in a significant number of false positives.

\subsection{Additional extended MI evaluation results}

To further broaden the scope of evaluation, we include additional results covering prompt refinement for ChatGPT-5, generic-domain setups, cross-MLLM consistency, label-only API-access MI, and cross-domain validation.

\subsubsection{Prompt refinement for ChatGPT-5 evaluation}

ChatGPT-5 can exhibit high refusal rates under prompts that directly mention identity-sensitive phrasing. We therefore evaluate a prompt variant with identity-related wording removed. The results in Tab.~\ref{tab:chatgpt5_prompt_refinement_supp} show that this refinement substantially reduces refusal rates while preserving strong discrimination between positive and negative pairs.

\begin{table*}[t]
\centering
\caption{ChatGPT-5 evaluation outcomes under original and identity-removed prompts.}
\begin{adjustbox}{width=0.8\textwidth}
\begin{tabular}{lcccccc}
\hline
\textbf{Prompt} & \textbf{Pos. Yes} & \textbf{Pos. No} & \textbf{Pos. Refusal} & \textbf{Neg. Yes} & \textbf{Neg. No} & \textbf{Neg. Refusal} \\ \hline
Original & 17.50\% & 2.67\% & 79.83\% & 0.09\% & 23.04\% & 76.86\% \\ \hline
Identity removal & 95.19\% & 2.87\% & 1.94\% & 5.57\% & 93.32\% & 1.11\% \\ \hline
\end{tabular}
\end{adjustbox}
\label{tab:chatgpt5_prompt_refinement_supp}
\end{table*}

\subsubsection{Agreement with human judgment on false positives}

To compare evaluator behavior against human judgment, we measure false-positive rates under $\mathcal{F}_{\text{MLLM}}$ and $\mathcal{F}_{\text{Curr}}$. As shown in Tab.~\ref{tab:fp_compare_human_supp}, $\mathcal{F}_{\text{MLLM}}$ yields substantially fewer false positives.

\begin{table}[t]
\centering
\caption{False-positive rates measured with human judgment as reference.}
\begin{tabular}{lc}
\hline
 & \textbf{FP rate} \\ \hline
$\mathcal{F}_{\text{MLLM}}$ & 15.13\% \\ \hline
$\mathcal{F}_{\text{Curr}}$ & 43.70\% \\ \hline
\end{tabular}
\label{tab:fp_compare_human_supp}
\end{table}

\subsubsection{Generic-domain setup: CIFAR-100}

Beyond face recognition, we evaluate a generic-domain setup by adapting PPA with $T$=ResNet18, $\mathcal{D}_{priv}$=CIFAR100, and $\mathcal{D}_{pub}$=CIFAR10. Results in Tab.~\ref{tab:cifar100_supp} show that $\mathcal{F}_{\text{Curr}}$ still exhibits a high FP rate.

\begin{table*}[t]
\centering
\caption{Results on the CIFAR-100 setup with adapted PPA.}
\begin{adjustbox}{width=0.7\textwidth}
\begin{tabular}{ccccccc}
\hline
\textbf{$\mathcal{F}_{\text{MLLM}}$} & \multicolumn{6}{c}{\textbf{$\mathcal{F}_{\text{Curr}}$}} \\ \hline
\textbf{AttAcc} & \textbf{$E$} & \textbf{AttAcc} & \textbf{FP rate} & \textbf{FN rate} & \textbf{TP rate} & \textbf{TN rate} \\ \hline
45.00\% & InceptionNetV3 & 66.00\% & 57.73\% & 23.89\% & 76.11\% & 42.27\% \\ \hline
\end{tabular}
\end{adjustbox}
\label{tab:cifar100_supp}
\end{table*}

\subsubsection{Consistency across eligible MLLMs}

We evaluate consistency across two eligible MLLMs. As shown in Tab.~\ref{tab:mllm_consistency_supp}, the evaluation outcomes are stable across model versions.

\begin{table}[t]
\centering
\caption{Evaluation outcomes across eligible MLLMs.}
\begin{tabular}{cc}
\hline
\textbf{Gemini 2.0} & \textbf{Gemini 2.5} \\ \hline
28.37\% & 29.08\% \\ \hline
\end{tabular}
\label{tab:mllm_consistency_supp}
\end{table}

In addition, sensitivity to the number of reference images remains low: varying from 2 to 5 reference images gives $29.16 \pm 1.82$, indicating stable evaluation behavior.

\subsubsection{Label-only API-access MI setup (BREP-MI)}

We also include a label-only API-access MI setup following BREP-MI with $T$=VGG16 and $\mathcal{D}_{priv}$=CelebA. Tab.~\ref{tab:brepmi_supp} again shows high FP rates under $\mathcal{F}_{\text{Curr}}$.

\begin{table*}[t]
\centering
\caption{Results on a label-only API-access BREP-MI setup.}
\begin{adjustbox}{width=0.66\textwidth}
\begin{tabular}{ccccccc}
\hline
\textbf{$\mathcal{F}_{\text{MLLM}}$} & \multicolumn{6}{c}{\textbf{$\mathcal{F}_{\text{Curr}}$}} \\ \hline
\textbf{AttAcc} & \textbf{$E$} & \textbf{AttAcc} & \textbf{FP rate} & \textbf{FN rate} & \textbf{TP rate} & \textbf{TN rate} \\ \hline
72.91\% & FaceNet112 & 80.94\% & 79.01\% & 18.35\% & 81.65\% & 20.99\% \\ \hline
\end{tabular}
\end{adjustbox}
\label{tab:brepmi_supp}
\end{table*}

\subsubsection{Cross-domain evaluator validation: StanfordDogs}

To validate the evaluator outside face recognition, we evaluate a dog-recognition setup on StanfordDogs. The results in Tab.~\ref{tab:stanforddogs_supp} show high yes/no agreement and zero refusal rates.

\begin{table*}[t]
\centering
\caption{Cross-domain evaluator validation on StanfordDogs.}
\begin{adjustbox}{width=0.82\textwidth}
\begin{tabular}{lcccccc}
\hline
\textbf{Prompt} & \textbf{Pos. Yes} & \textbf{Pos. No} & \textbf{Pos. Refusal} & \textbf{Neg. Yes} & \textbf{Neg. No} & \textbf{Neg. Refusal} \\ \hline
Original prompt & 90.08\% & 9.92\% & 0.00\% & 1.25\% & 98.75\% & 0.00\% \\ \hline
\end{tabular}
\end{adjustbox}
\label{tab:stanforddogs_supp}
\end{table*}

\begin{figure*}[t]
  \centering
  \begin{adjustbox}{width=0.8\textwidth,center}
  \includegraphics[width=\textwidth]{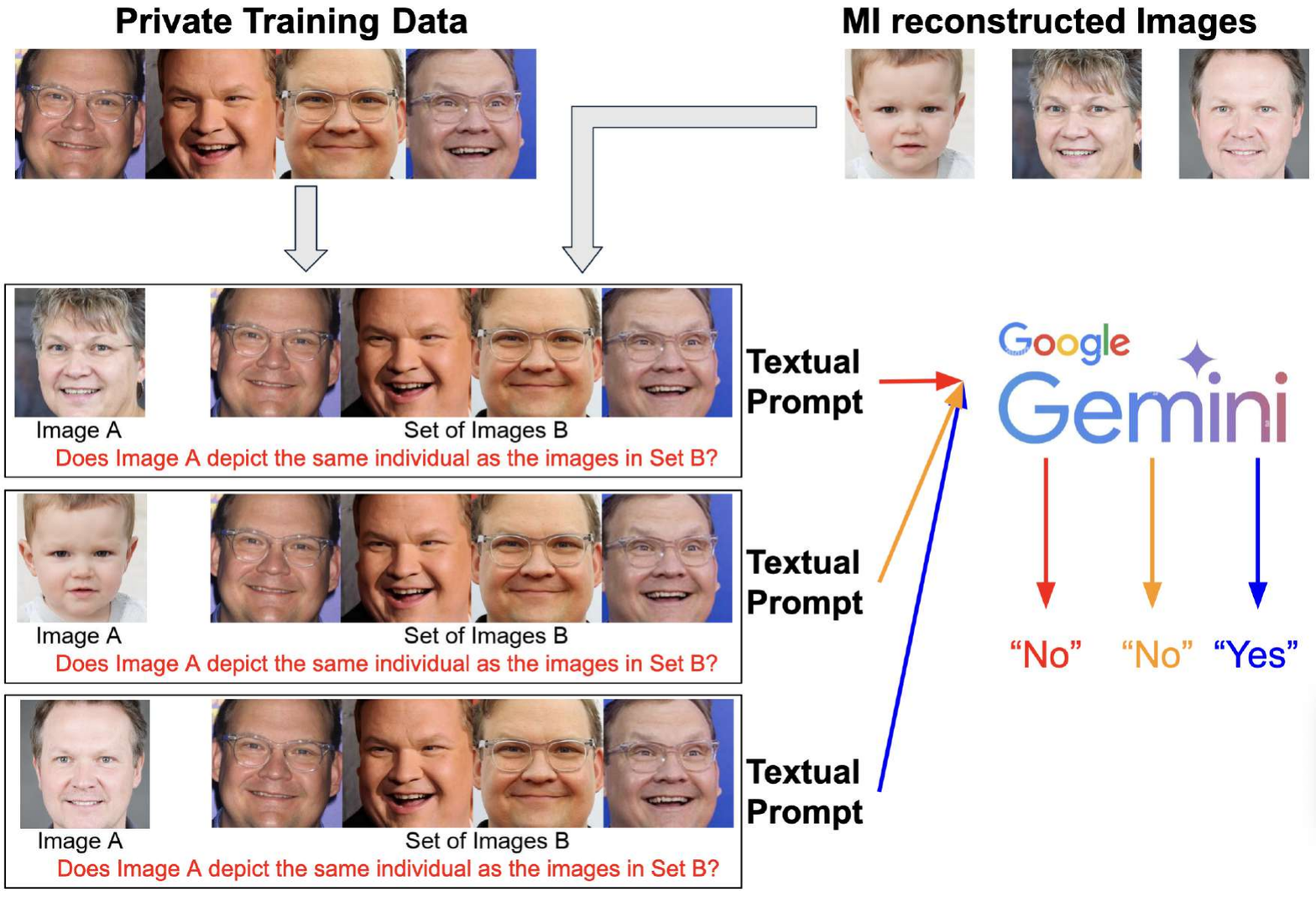}
  \end{adjustbox}
  \caption{\textbf{Our detailed implementation of MLLM-based MI Evaluation Framework $\boldmath{\mathcal{F}_{MLLM}}$}. For each reconstructed image, we pair with a set of private training data to construct an evaluation query image. Then, each evaluation query image is passed to Gemini with a textual prompt. The detailed of textual prompt can be found in Sec.~\ref{sec: The detailed prompt}. The final attack accuracy is computed based on Gemini's responses.}
  \label{fig:F_MLLM}
\end{figure*}

\section{Detailed experimental reproducibility}

\subsection{Detailed Implementation} \label{Sec: F_MLLM}

Our implementation of $\mathcal{F}_{MLLM}$ is illustrated in Fig.~\ref{fig:F_MLLM}. To evaluate whether a reconstructed image is a successful or unsuccessful attack, we employ Gemini 2.0 Flash API (see the main manuscript for our justification for choosing Gemini) for the evaluation.

Given a reconstructed image (Image A), we construct an evaluation query image by pairing it with a set of private training images (Set B) that includes the target identity. We then formulate a natural language textual prompt along with the evaluation query image and pass it to Gemini. The textual prompts are shown in the table below and are fixed across evaluation queries for a fair comparison.

\begin{table*}[ht]
\Large
\setlength{\tabcolsep}{8pt}
\centering
\caption{\textbf{Our controlled experiment to show the effect of Type I adversarial attacks in MI on false positive rates.} We provide results on this experiment on additional setups in Appx.~\ref{Sec:Detailed MI setup}}

\begin{adjustbox}{width=0.7\textwidth}
\begin{tabular}{ccccccc}
\hline
\textbf{Attack}         & \textbf{$E$}                  & \textbf{$\mathcal{D}_{priv}$} & \textbf{$\mathcal{D}_{pub}$} & \textbf{$T$}                 & \multicolumn{1}{l}{} & \textbf{\makecell{FP rates under $E$}} \\ \hline
\multirow{8}{*}{PPA}    & \multirow{10}{*}{InceptionV3} & \multirow{10}{*}{Facescrub}   & \multirow{8}{*}{FFHQ}        & \multirow{2}{*}{Resnet101}   & Neg $x^{r}_y$        & 82.71\%                     \\
                        &                               &                               &                              &                              & Neg $x^{natural}_y$  & 0.94\%                      \\ \cline{5-7} 
                        &                               &                               &                              & \multirow{2}{*}{Resnet152}   & Neg $x^{r}_y$        & 85.09\%                     \\
                        &                               &                               &                              &                              & Neg $x^{natural}_y$  & 0.94\%                      \\ \cline{5-7} 
                        &                               &                               &                              & \multirow{2}{*}{MaxViT}      & Neg $x^{r}_y$        & 79.48\%                     \\
                        &                               &                               &                              &                              & Neg $x^{natural}_y$  & 0.94\%                      \\ \cline{5-7} 
                        &                               &                               &                              & \multirow{2}{*}{DenseNet121} & Neg $x^{r}_y$        & 72.41\%                     \\
                        &                               &                               &                              &                              & Neg $x^{natural}_y$  & 0.94\%                      \\ \cline{1-1} \cline{4-7} 
\multirow{2}{*}{IFGMI}  &                               &                               & \multirow{2}{*}{MetFaces}    & \multirow{2}{*}{Resnet18}    & Neg $x^{r}_y$        & 72.71\%                     \\
                        &                               &                               &                              &                              & Neg $x^{natural}_y$  & 0.94\%                      \\ \hline
\multirow{2}{*}{PLGMI}  & \multirow{26}{*}{FaceNet112}  & \multirow{26}{*}{CelebA}      & \multirow{2}{*}{FFHQ}        & \multirow{2}{*}{VGG16}       & Neg $x^{r}_y$        & 88.49\%                     \\
                        &                               &                               &                              &                              & Neg $x^{natural}_y$  & 0.00\%                      \\ \cline{1-1} \cline{4-7} 
\multirow{12}{*}{LOMMA} &                               &                               & \multirow{6}{*}{CelebA}      & \multirow{2}{*}{FaceNet64}   & Neg $x^{r}_y$        & 93.73\%                     \\
                        &                               &                               &                              &                              & Neg $x^{natural}_y$  & 0.00\%                      \\ \cline{5-7} 
                        &                               &                               &                              & \multirow{2}{*}{IR152}       & Neg $x^{r}_y$        & 86.80\%                     \\
                        &                               &                               &                              &                              & Neg $x^{natural}_y$  & 0.00\%                      \\ \cline{5-7} 
                        &                               &                               &                              & \multirow{2}{*}{VGG16}       & Neg $x^{r}_y$        & 90.70\%                     \\
                        &                               &                               &                              &                              & Neg $x^{natural}_y$  & 0.00\%                      \\ \cline{4-7} 
                        &                               &                               & \multirow{6}{*}{FFHQ}        & \multirow{2}{*}{FaceNet64}   & Neg $x^{r}_y$        & 69.73\%                     \\
                        &                               &                               &                              &                              & Neg $x^{natural}_y$  & 0.00\%                      \\ \cline{5-7} 
                        &                               &                               &                              & \multirow{2}{*}{IR152}       & Neg $x^{r}_y$        & 77.85\%                     \\
                        &                               &                               &                              &                              & Neg $x^{natural}_y$  & 0.00\%                      \\ \cline{5-7} 
                        &                               &                               &                              & \multirow{2}{*}{VGG16}       & Neg $x^{r}_y$        & 61.55\%                     \\
                        &                               &                               &                              &                              & Neg $x^{natural}_y$  & 0.00\%                      \\ \cline{1-1} \cline{4-7} 
\multirow{12}{*}{KEDMI} &                               &                               & \multirow{6}{*}{CelebA}      & \multirow{2}{*}{FaceNet64}   & Neg $x^{r}_y$        & 78.40\%                     \\
                        &                               &                               &                              &                              & Neg $x^{natural}_y$  & 0.00\%                      \\ \cline{5-7} 
                        &                               &                               &                              & \multirow{2}{*}{IR152}       & Neg $x^{r}_y$        & 74.55\%                     \\
                        &                               &                               &                              &                              & Neg $x^{natural}_y$  & 0.00\%                      \\ \cline{5-7} 
                        &                               &                               &                              & \multirow{2}{*}{VGG16}       & Neg $x^{r}_y$        & 69.80\%                     \\
                        &                               &                               &                              &                              & Neg $x^{natural}_y$  & 0.00\%                      \\ \cline{4-7} 
                        &                               &                               & \multirow{6}{*}{FFHQ}        & \multirow{2}{*}{FaceNet64}   & Neg $x^{r}_y$        & 52.24\%                     \\
                        &                               &                               &                              &                              & Neg $x^{natural}_y$  & 0.00\%                      \\ \cline{5-7} 
                        &                               &                               &                              & \multirow{2}{*}{IR152}       & Neg $x^{r}_y$        & 51.02\%                     \\
                        &                               &                               &                              &                              & Neg $x^{natural}_y$  & 0.00\%                      \\ \cline{5-7} 
                        &                               &                               &                              & \multirow{2}{*}{VGG16}       & Neg $x^{r}_y$        & 41.33\%                     \\
                        &                               &                               &                              &                              & Neg $x^{natural}_y$  & 0.00\%                      \\ \hline
\end{tabular}
\end{adjustbox}

\label{tab:W-Wo-Adv-Investigation}
\end{table*}

For each reconstructed image, the model outputs a categorical response (``Yes'' or ``No''). A ``Yes'' answer is interpreted as a successful attack. By evaluating a large number of such queries and computing the proportion of correct identifications, $\mathcal{F}_{MLLM}$ provide an automated and faithful evaluation of MI.

\subsection{The detailed prompt}
\label{sec: The detailed prompt}

The detailed textual prompts in our MI evaluation framework can be found in Tab. \ref{tab:prompt}.

\begin{table*}[h]
\Large
\centering

\caption{Detailed textual prompts in our MI evaluation framework $\mathcal{F}_{MLLM}$}
\begin{adjustbox}{width=2.0\columnwidth}
\begin{tabular}{ccc}
\hline
Domain           & $\mathcal{D}_{priv}$ & Textual Prompt                                                                                                                                                                                                                          \\ \hline
\multirow{2}{*}{Face Recognition} & Facescrub            & \multirow{2}{*}{\makecell{You are an expert in face recognition. Taking into account the face aging, lighting, different hair styles, \\wearing and not wearing of eye glasses or other accessory, do the task in the image. Only answer yes or no}} \\ \cline{2-2}
                 & CelebA               &                                                                                                                                                                                                                                         \\ \hline
Dog Recognition  & Stanford Dogs        & \makecell{You are an expert in dog breed recognition. Taking into account the lighting and viewpoint, \\ do the task in the image. Only answer yes or no}                                                                                             \\ \hline
\end{tabular}
\end{adjustbox}

\label{tab:prompt}
\end{table*}

\subsection{Error Bar of Evaluation results}

As mentioned in the main manuscript, we provide an error bar of evaluation results with $\mathcal{F}_{MLLM}$ to further demonstrate the robustness of our proposed MI evaluation framework. The results can be found in Tab.~\ref{tab:error_bar}

\begin{table*}[t]
\large
\renewcommand{\arraystretch}{1.5}
\centering
\vspace{-0.3cm}
\caption{{\bf Our investigation on MI evaluation framework using our comprehensive dataset of MI attack samples.} We run the evaluations with our $\mathcal{F}_{MLLM}$ threes times and report mean $\pm$ std.}
\begin{adjustbox}{width=1.9\columnwidth}
\begin{tabular}{ccc
>{\columncolor[HTML]{E9F1E6}}c 
>{\columncolor[HTML]{EFEAEA}}c 
>{\columncolor[HTML]{EFEAEA}}c 
>{\columncolor[HTML]{EFEAEA}}c 
>{\columncolor[HTML]{EFEAEA}}c 
>{\columncolor[HTML]{EFEAEA}}c 
>{\columncolor[HTML]{EFEAEA}}c 
>{\columncolor[HTML]{EFEAEA}}c }
\hline
                                     &                                  &                                 & \cellcolor[HTML]{E9F1E6}                             & \textbf{$\mathcal{F}_{MLLM}$} & \multicolumn{6}{c}{\cellcolor[HTML]{EFEAEA}\textbf{$\mathcal{F}_{Curr}$}}                                                                                           \\ \cline{6-11} 
\multirow{-2}{*}{\textbf{MI Attack}} & \multirow{-2}{*}{\textbf{$\mathcal{D}_{pub}$}} & \multirow{-2}{*}{\textbf{$\mathcal{D}_{pub}$}} & \multirow{-2}{*}{\cellcolor[HTML]{E9F1E6}\textbf{$T$}} & \textbf{AttAcc}   & \textbf{$E$}                                               & \textbf{AttAcc} & \textbf{FP rate} & \textbf{FN rate} & \textbf{TP rate} & \textbf{TN rate} \\ \hline
                                     &                                  &                                 & ResNet18                                             & 28.22$\pm$0.30\%            & \cellcolor[HTML]{EFEAEA}                                 &   91.39\%       & 90.03$\pm$0.09\%          & 4.82$\pm$0.51\%           & 94.82$\pm$0.32\%          & 9.97$\pm$0.09\%           \\ \cline{4-5} \cline{7-11} 
                                     &                                  &                                 & ResNet101                                            & 28.48$\pm$0.36\%            & \cellcolor[HTML]{EFEAEA}                                 & 84.69\%         & 82.79$\pm$0.09\%          & 10.52$\pm$0.16\%          & 89.48$\pm$0.16\%          & 17.21$\pm$0.09\%          \\ \cline{4-5} \cline{7-11} 
                                     &                                  &                                 & ResNet152                                            & 30.20$\pm$0.09\%            & \cellcolor[HTML]{EFEAEA}                                 & 86.84\%         & 85.13$\pm$0.14\%          & 9.21$\pm$0.34\%           & 90.79$\pm$0.34\%          & 14.87$\pm$0.14\%          \\ \cline{4-5} \cline{7-11} 
                                     &                                  &                                 & DenseNet121                                          & 27.44$\pm$0.27\%            & \cellcolor[HTML]{EFEAEA}                                 & 72.41\%         & 70.11$\pm$0.05\%          & 21.52$\pm$0.07\%          & 78.48$\pm$0.07\%          & 29.89$\pm$0.05\%          \\ \cline{4-5} \cline{7-11} 
   \multirow{-5}{*}{PPA}                                  & \multirow{-5}{*}{FaceScrub}      & \multirow{-5}{*}{FFHQ}          & MaxViT                                               & 30.30$\pm$0.13\%            & \multirow{-5}{*}{\cellcolor[HTML]{EFEAEA}InceptionNetV3} & 79.48\%         & 77.32$\pm$0.16\%          & 15.54$\pm$0.33\%          & 84.46$\pm$0.34\%          & 22.68$\pm$0.14\%          \\ \cline{1-11} 
                                     &                                  & FFHQ                            & ResNet18                                             & 34.14$\pm$0.29\%            & \cellcolor[HTML]{EFEAEA}                                 & 95.85\%         & 94.61$\pm$0.03\%          & 1.75$\pm$0.07\%           & 98.25$\pm$0.07\%          & 5.39$\pm$0.03\%           \\ \cline{3-5} \cline{7-11} 
\multirow{-2}{*}{IFGMI}              & \multirow{-2}{*}{FaceScrub }      & Metfaces                         & ResNet18                                             & 1.57$\pm$0.07\%             & \multirow{-2}{*}{\cellcolor[HTML]{EFEAEA}InceptionNetV3} & 72.50\%         & 72.24$\pm$0.05\%          & 11.39$\pm$3.74\%           & 88.61$\pm$3.74\%          & 27.76$\pm$0.05\%          \\ \hline
                                     &                                  & CelebA                          & VGG16                                                 & 73.51$\pm$0.75\%            & \cellcolor[HTML]{EFEAEA}                                 & 98.73\%         & 99.33$\pm$0.14\%          & 1.48$\pm$0.05\%           & 98.52$\pm$0.05\%          & 0.67$\pm$0.14\%           \\ \cline{3-5} \cline{7-11} 
\multirow{-2}{*}{PLGMI}              & \multirow{-2}{*}{CelebA }         & FFHQ                            & VGG16                                                & 48.59$\pm$0.57\%            & \multirow{-2}{*}{\cellcolor[HTML]{EFEAEA}FaceNet112}     & 88.67\%         & 88.51$\pm$0.07\%          & 11.16$\pm$0.06\%          & 88.84$\pm$0.06\%          & 11.47$\pm$0.07\%          \\ \hline
                                     &                                  &                                 & IR152                                                & 79.02$\pm$0.30\%            & \cellcolor[HTML]{EFEAEA}                                 & 92.00\%         & 92.47$\pm$1.22\%          & 8.13$\pm$0.33\%           & 91.87$\pm$0.33\%          & 7.53$\pm$1.22\%          \\ \cline{4-5} \cline{7-11} 
                                     &                                  &                                 & FaceNet64                                            & 79.76$\pm$0.27\%            & \cellcolor[HTML]{EFEAEA}                                 & 90.40\%         & 87.71$\pm$0.80\%          & 8.92$\pm$0.20\%           & 91.08$\pm$0.20\%          & 12.29$\pm$0.08\%           \\ \cline{4-5} \cline{7-11} 
                                     &                                  & \multirow{-3}{*}{CelebA}        & VGG16                                                & 80.73$\pm$0.77\%            & \cellcolor[HTML]{EFEAEA}                                 & 90.13\%         & 90.29$\pm$0.95\%          & 9.91$\pm$0.22\%          & 90.09$\pm$0.21\%          & 9.71$\pm$0.95\%           \\ \cline{3-5} \cline{7-11} 
                                     &                                  &                                 & IR152                                                & 45.60$\pm$0.58\%            & \cellcolor[HTML]{EFEAEA}                                 & 77.73\%         & 77.37$\pm$0.45\%          & 21.84$\pm$0.52\%          & 78.16$\pm$0.52\%          & 22.63$\pm$0.45\%          \\ \cline{4-5} \cline{7-11} 
                                     &                                  &                                 & FaceNet64                                            & 45.49$\pm$0.74\%            & \cellcolor[HTML]{EFEAEA}                                 & 72.13\%         & 69.92$\pm$0.18\%          & 25.21$\pm$0.17\%          & 74.79$\pm$0.17\%          & 30.08$\pm$0.18\%          \\ \cline{4-5} \cline{7-11} 
\multirow{-6}{*}{LOMMA}              & \multirow{-6}{*}{CelebA}         & \multirow{-3}{*}{FFHQ}          & VGG16                                                & 56.09$\pm$1.20\%            & \multirow{-6}{*}{\cellcolor[HTML]{EFEAEA}FaceNet112}     & 63.07\%         & 61.33$\pm$0.31\%          & 35.58$\pm$0.18\%          & 64.42$\pm$0.18\%          & 38.67$\pm$0.31\%          \\ \hline
                                     &                                  &                                 & IR152                                                & 67.24$\pm$0.83\%            & \cellcolor[HTML]{EFEAEA}                                 & 79.27\%         & 74.97$\pm$0.37\%          & 18.64$\pm$0.25\%          & 81.36$\pm$0.25\%          & 24.70$\pm$0.23\%          \\ \cline{4-5} \cline{7-11} 
                                     &                                  &                                 & FaceNet64                                            & 66.15$\pm$0.73\%            & \cellcolor[HTML]{EFEAEA}                                 & 80.53\%         & 77.27$\pm$1.04\%          & 17.81$\pm$0.49\%          & 82.19$\pm$0.49\%          & 30.15$\pm$1.56\%          \\ \cline{4-5} \cline{7-11} 
                                     &                                  & \multirow{-3}{*}{CelebA}        & VGG16                                                & 69.38$\pm$1.04\%            & \cellcolor[HTML]{EFEAEA}                                 & 73.13\%         & 69.85$\pm$1.56\%          & 25.44$\pm$0.62\%          & 74.56$\pm$0.62\%          & 30.15$\pm$1.56\%          \\ \cline{3-5} \cline{7-11} 
                                     &                                  &                                 & IR152                                                & 36.96$\pm$0.62\%            & \cellcolor[HTML]{EFEAEA}                                 & 52.20\%         & 50.24$\pm$0.75\%          & 44.42$\pm$1.34\%          & 55.58$\pm$1.34\%          & 49.76$\pm$0.75\%          \\ \cline{4-5} \cline{7-11} 
                                     &                                  &                                 & FaceNet64                                            & 35.96$\pm$0.14\%            & \cellcolor[HTML]{EFEAEA}                                 & 54.60\%         & 52.08$\pm$0.62\%          & 40.91$\pm$1.13\%          & 59.09$\pm$1.13\%          & 47.92$\pm$0.62\%          \\ \cline{4-5} \cline{7-11} 
\multirow{-6}{*}{KEDMI}              & \multirow{-6}{*}{CelebA}         & \multirow{-3}{*}{FFHQ}          & VGG16                                                & 38.85$\pm$0.80\%            & \multirow{-6}{*}{\cellcolor[HTML]{EFEAEA}FaceNet112}     & 42.47\%         & 41.24$\pm$0.36\%          & 55.50$\pm$0.58\%          & 44.40$\pm$0.58\%          & 58.76$\pm$0.36\%          \\ \hline
\end{tabular}
\end{adjustbox}

\vspace{-0.5cm}
\label{tab:error_bar}
\end{table*}

\subsection{Detailed MI setup} \label{Sec:Detailed MI setup}

To ensure the reproducibility, we strictly follow previous studies \citep{zhang2020secret,chen2021knowledge,nguyen_2023_CVPR,struppek2022plug,qiu2024closer,koh2024vulnerability,ho2024model} for MI setups.

\paragraph{MI attacks.} Our study focuses on SOTA GAN-based MI attack that achieve strong performance in computer vision domain. These attacks optimize the GAN latent space rather than directly optimize the image space.

\textit{KEDMI} \citep{chen2021knowledge} Introduces an MI-specific GAN that incorporates knowledge from the target classifier. The discriminator performs dual tasks: distinguishing real and fake samples and predicting class-wise labels.

\textit{LOMMA} \citep{nguyen_2023_CVPR} Improves MI attacks using a novel logit loss and model augmentation to mitigate overfitting.

\textit{PLGMI} \citep{yuan2023pseudo} Leverages conditional GANs to isolate class-specific search spaces and uses Max-Margin Loss to address vanishing gradients in MI optimization.

\textit{PPA} \citep{struppek2022plug} Utilizes powerful StyleGAN for high-resolution image MI attacks, emphasizing a modular design adaptable to different architectures and datasets.

\textit{IFGMI} \citep{qiu2024closer} Proposes Intermediate Features Generative Model Inversion, extending optimization from latent codes to intermediate features, enhancing the attack's expressive capability.

\paragraph{MI defense.} Our study focuses on SOTA MI defenses. Differ from MI attacks, MI defenses aim to minimize the disclosure of training samples during the MI optimization process. 

\textit{TL \citep{ho2024model}} Leverages Transfer Learning to limit sensitive information encoding in earlier layers, degrading MI attack performance.

\textit{LS \citep{struppek2023careful}} Introduces label smoothing with negative factors to impede class-related information extraction.

\textit{RoLSS \citep{koh2024vulnerability}} Demonstrates that removing skip connections in the last stage significantly reduces MI attack accuracy, offering a better MI robustness trade-off.

\textit{TTS \citep{koh2024vulnerability}} Buliding on top of RoLSS. Particularly, in the first stage, the model $T$ with full skip-connections architecture is trained on private dataset. Then in the stage 2, the skip connection removed architecture, i.e. RoLSS, is fine-tuned on private dataset. The pre-trained parameters in Stage 1 serves as initialization for the stage 2, thereby improve the convergence of model in stage 2. 

\paragraph{Private training data \boldmath{$\mathcal{D}_{priv}$}.} Following previous works \citep{zhang2020secret,chen2021knowledge,nguyen_2023_CVPR,struppek2022plug,qiu2024closer,koh2024vulnerability,ho2024model}, we focus on reconstruction of images and use the
face recognition as a running example including FaceScrub \citep{ng2014data} and CelebA \citep{liu2015deep}. 

\textit{FaceScrub} \citep{ng2014data}: FaceScrub provides cropped facial images for 530 identities. The dataset publicly a total of 37,878 images. After train/test splitting, this resulted in 34,090 training samples and 3,788 test samples.

\begin{figure}[t]
  \centering
  \begin{adjustbox}{width=0.65\textwidth,center}
  \includegraphics[width=\textwidth]{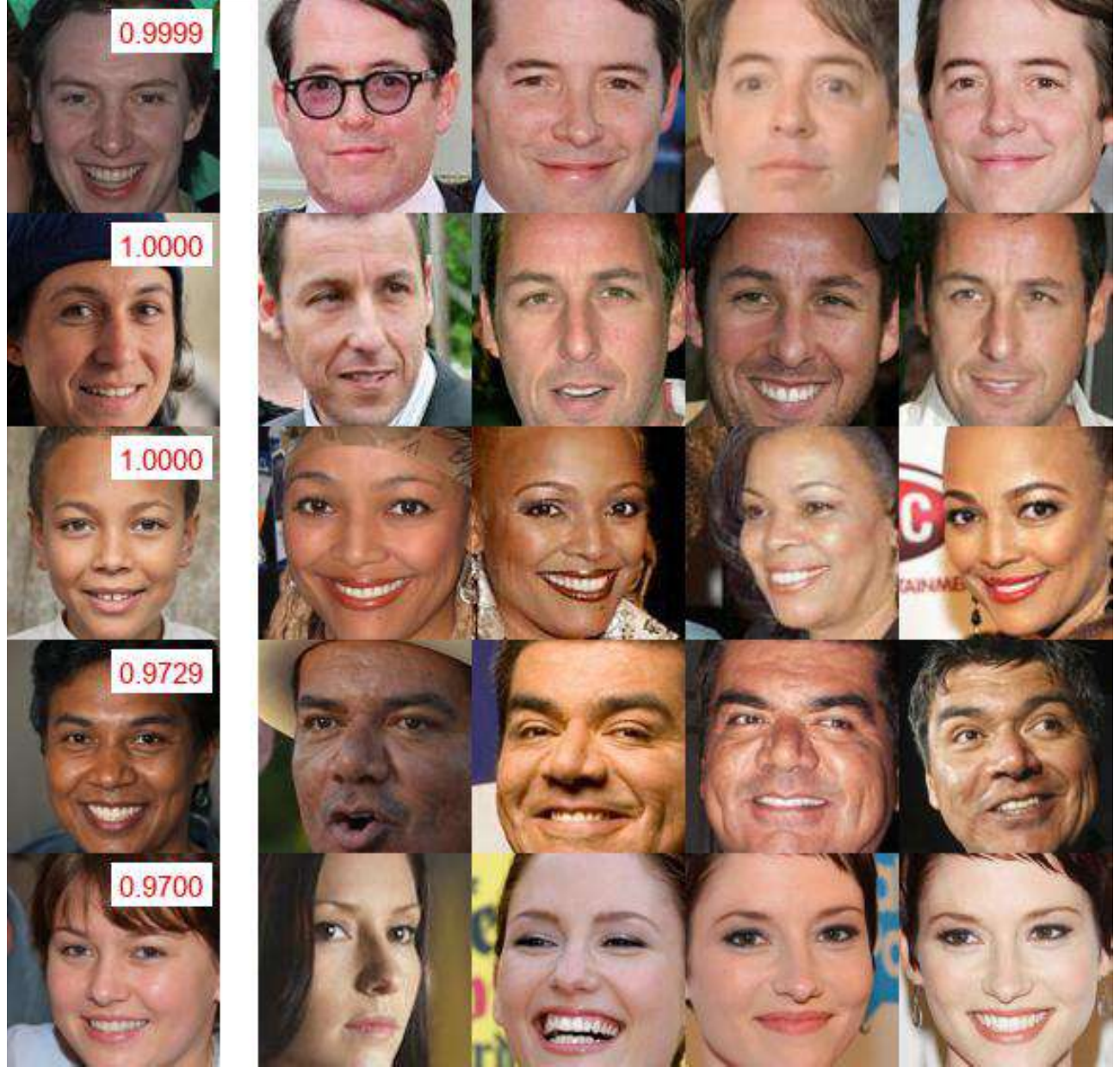}
  \end{adjustbox}
  \caption{Additional visualization of false positives. These MI false positives do not capture visual identity features of the target individual in the private training data, but they are still deemed successful attacks according to $\mathcal{F}_{Curr}$ with a high confidence (indicated in red text). Here, $T$=MaxViT \citep{tu2022maxvit}, $\mathcal{D}_{priv}$=FaceScrub \citep{ng2014data}, $\mathcal{D}_{pub}$=FFHQ \citep{karras2019style}, $E$=InceptioNetV3 under PPA attack \citep{struppek2022plug}.}
  \label{fig:fp-maxvit-ppa}
\end{figure}

\begin{figure}[h]
  \centering
  \begin{adjustbox}{width=0.65\textwidth,center}
  \includegraphics[width=\textwidth]{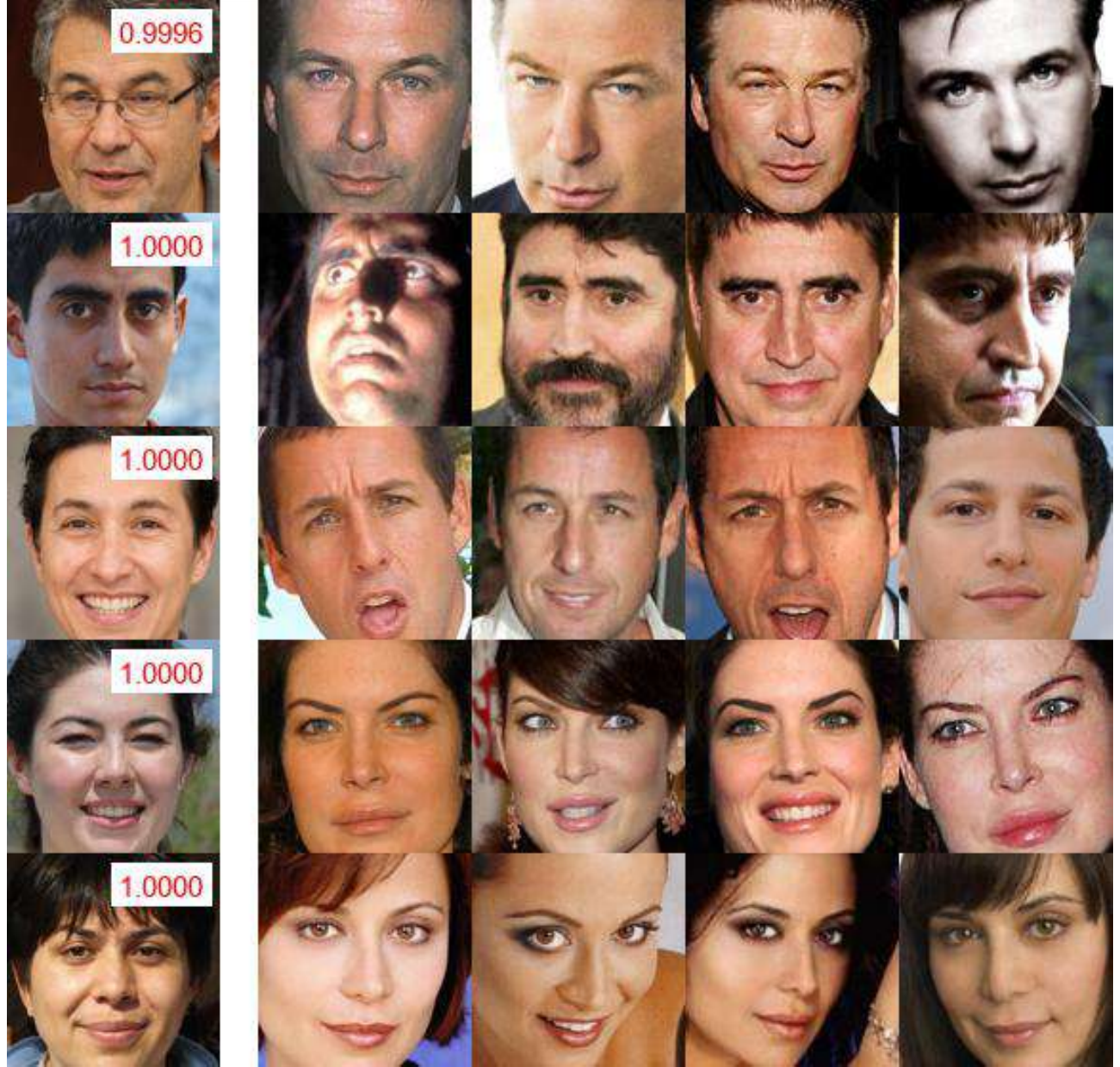}
  \end{adjustbox}
  \caption{Additional visualization of false positives. These MI false positives do not capture visual identity features of the target individual in the private training data, but they are still deemed successful attacks according to $\mathcal{F}_{Curr}$ with a high confidence (indicated in red text). Here, $T$=DenseNet121 \citep{huang2017densely}, $\mathcal{D}_{priv}$=FaceScrub \citep{ng2014data}, $\mathcal{D}_{pub}$=FFHQ \citep{karras2019style}, $E$=InceptioNetV3 under PPA attack \citep{struppek2022plug}.}
  \label{fig:fp-d121-ppa}
\end{figure}

\begin{figure}[h]
  \centering
  \begin{adjustbox}{width=0.65\textwidth,center}
  \includegraphics[width=\textwidth]{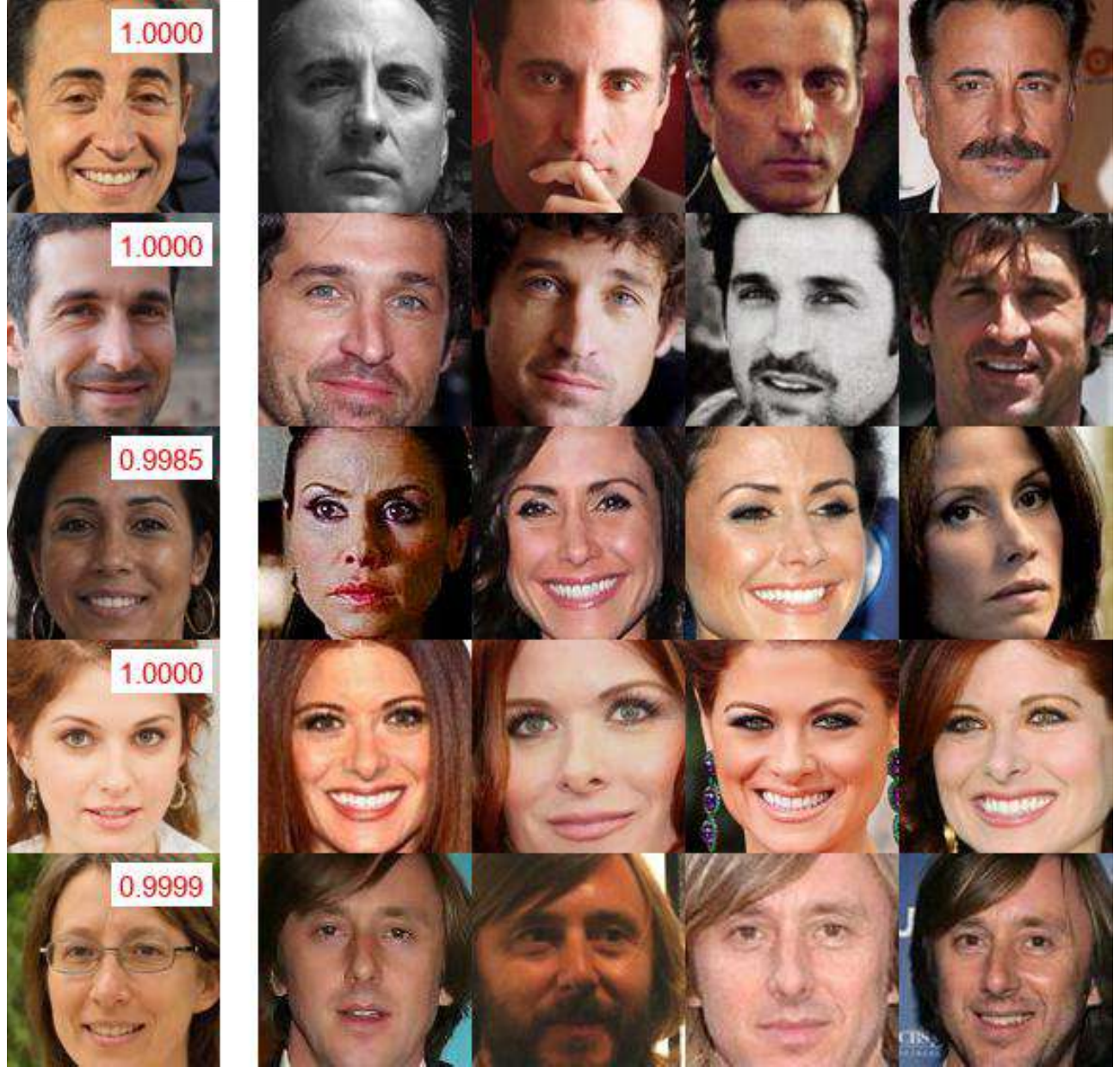}
  \end{adjustbox}
  \caption{Additional visualization of false positives. These MI false positives do not capture visual identity features of the target individual in the private training data, but they are still deemed successful attacks according to $\mathcal{F}_{Curr}$ with a high confidence (indicated in red text). Here, $T$=ResNet101 \citep{he2016deep}, $\mathcal{D}_{priv}$=FaceScrub \citep{ng2014data}, $\mathcal{D}_{pub}$=FFHQ \citep{karras2019style}, $E$=InceptioNetV3 under PPA attack \citep{struppek2022plug}.}
  \label{fig:fp-r101-ppa}
\end{figure}

\begin{figure}[h]
  \centering
  \begin{adjustbox}{width=0.65\textwidth,center}
  \includegraphics[width=\textwidth]{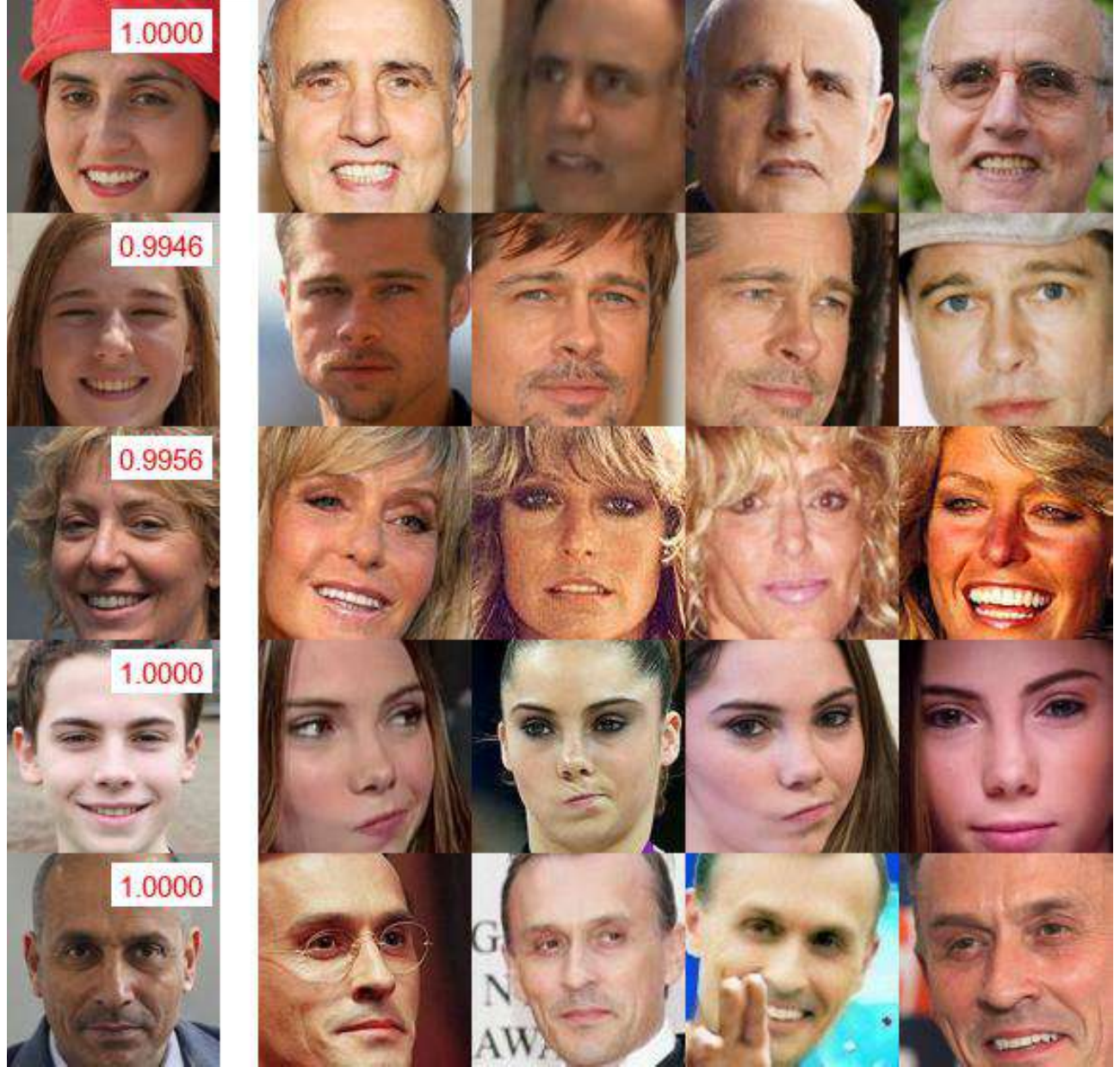}
  \end{adjustbox}
  \caption{Additional visualization of false positives. These MI false positives do not capture visual identity features of the target individual in the private training data, but they are still deemed successful attacks according to $\mathcal{F}_{Curr}$ with a high confidence (indicated in red text). Here, $T$=ResNet152 \citep{he2016deep}, $\mathcal{D}_{priv}$=FaceScrub \citep{ng2014data}, $\mathcal{D}_{pub}$=FFHQ \citep{karras2019style}, $E$=InceptioNetV3 under PPA attack \citep{struppek2022plug}.}
  \label{fig:fp-r152-ppa}
\end{figure}

\begin{figure}[h]
  \centering
  \begin{adjustbox}{width=0.65\textwidth,center}
  \includegraphics[width=\textwidth]{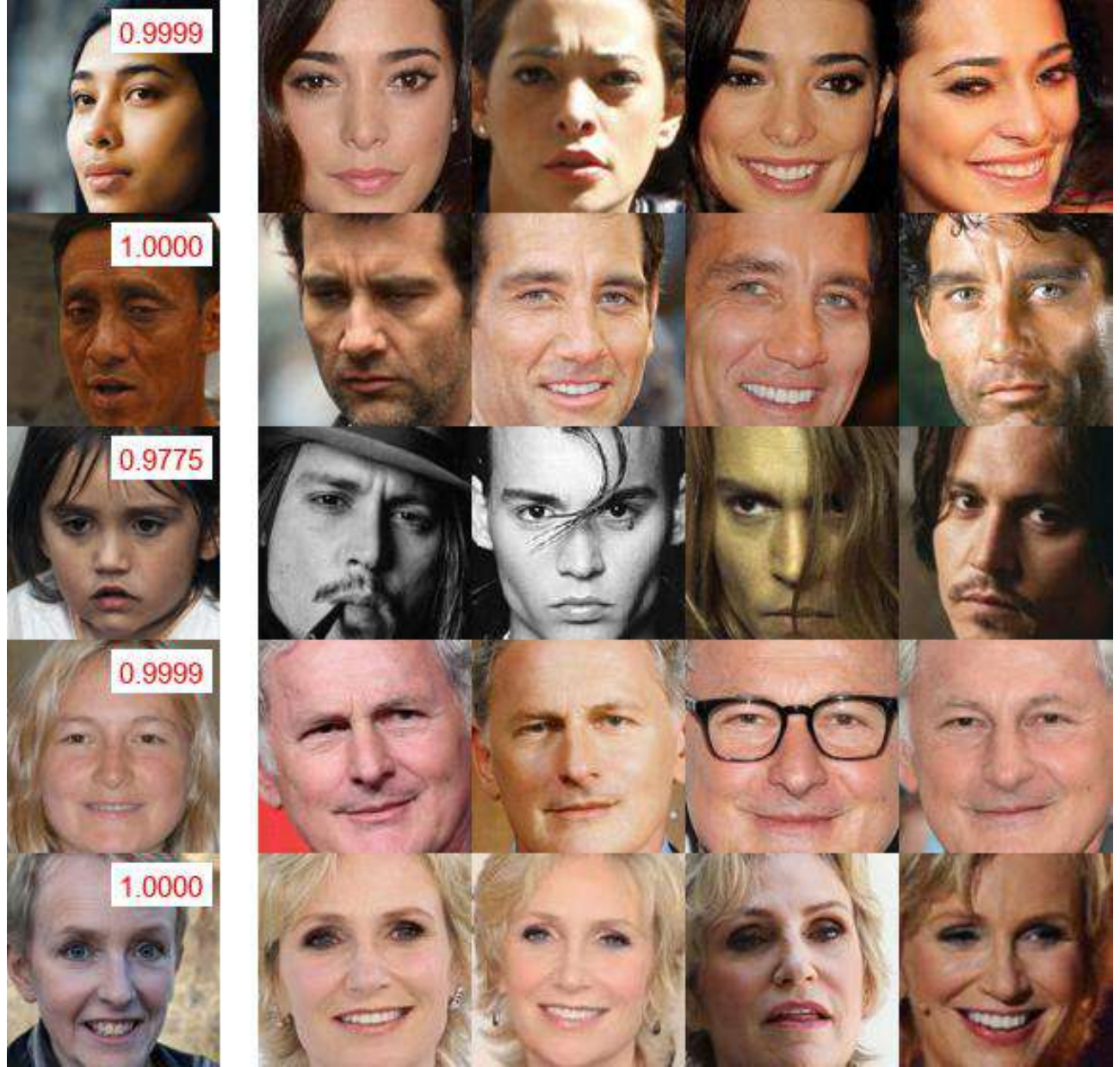}
  \end{adjustbox}
  \caption{Additional visualization of false positives. These MI false positives do not capture visual identity features of the target individual in the private training data, but they are still deemed successful attacks according to $\mathcal{F}_{Curr}$ with a high confidence (indicated in red text). Here, $T$=ResNet18 \citep{he2016deep}, $\mathcal{D}_{priv}$=FaceScrub \citep{ng2014data}, $\mathcal{D}_{pub}$=FFHQ \citep{karras2019style}, $E$=InceptioNetV3 under IFGMI attack \citep{qiu2024closer}.}
  \label{fig:fp-r18-ifgmi}
\end{figure}

\textit{CelebA} \citep{liu2015deep}: CelebA is a dataset of celebrity facial images available for non-commercial research. Following previous works \citep{zhang2020secret,chen2021knowledge,nguyen_2023_CVPR,struppek2022plug,qiu2024closer,koh2024vulnerability,ho2024model}, we select the top 1,000 identities with the most samples from 10,177 available identities, resulting in 27,034 training samples and 3,004 test samples.

\paragraph{Public data for GAN \boldmath{$\mathcal{D}_{pub}$}.} Following the data preparation in previous
works \citep{zhang2020secret,chen2021knowledge,nguyen_2023_CVPR,struppek2022plug,qiu2024closer,koh2024vulnerability,ho2024model}, we use $\mathcal{D}_{pub}$ ensuring that the dataset
$\mathcal{D}_{priv}$ and $\mathcal{D}_{pub}$ with no class
intersection. $\mathcal{D}_{priv}$ is used to train the target classifier $T$,
while $\mathcal{D}_{pub}$ is used to train GAN to extract general features only.

\textit{CelebA} \citep{liu2015deep}: Following previous works \citep{zhang2020secret,chen2021knowledge,nguyen_2023_CVPR,ho2024model}, we select 30,000 images from identities distinct from the 1,000 identities in $\mathcal{D}_{priv}$.

\textit{FFHQ} \citep{karras2019style}: This dataset contains 70,000 high-quality human face images sourced from Flickr, offering significant diversity in age, ethnicity, and backgrounds.

\textit{MetFaces \citep{karras2020training}}. This dataset includes 1,336 high-quality artistic renderings of human faces, covering various art styles. The images exhibit significant diversity and uniqueness.

\paragraph{Target Classifier \boldmath{$T$}.} Following previous works \citep{zhang2020secret,chen2021knowledge,nguyen_2023_CVPR,struppek2022plug,qiu2024closer,koh2024vulnerability,ho2024model}, we include a wide ranges of architectures as $T$ in our study including ResNet18/101/152 \citep{he2016deep}, DenseNet121\citep{huang2017densely}, MaxViT \citep{tu2022maxvit}, FaceNet \citep{chen2021knowledge}, and VGG16 \citep{he2016deep}. To ensure the reproducibility, we utilize the checkpoints of these target classifier in the previous works.

\subsection{Computing resources}

We conducted all experiments on NVIDIA RTX A5000 GPUs running Ubuntu 20.04.2 LTS, with AMD Ryzen Threadripper PRO 5975WX 32-Core processors. The environment setup includes CUDA 12.2, Python 3.8.18, and PyTorch 1.12.0 with Torchvision 0.14.1. For high-resolution tasks, \citep{struppek2022plug,qiu2024closer}, we use model architectures and pre-trained ImageNet backbone weights from Torchvision. For the low-resolution setup, following \citep{chen2021knowledge,nguyen_2023_CVPR,ho2024model}, we employed VGG architecture with pre-trained ImageNet weights from Torchvision, while we utilize IR152 and FaceNet architectures with pre-trained backbones from face.evoLVe\footnote{https://github.com/ZhaoJ9014/face.evoLVe}.

We employ the Gemini 2.0 Flash API in $\mathcal{F}_{\text{MLLM}}$ and emphasize that our implementation is both reliable and cost efficient. Particularly, in our implementation, each evaluation query costs \$0.0002886 (see the official Gemini API documentation\footnote{\url{https://ai.google.dev/gemini-api/docs}} for cost estimation). This cost is reasonable for large-scale evaluations. For example, in our study involving larger-scale 26 experimental setups and a total of 71,880 MI-reconstructed images, the overall cost is around \$20.75, making our evaluation framework scalable and accessible for future research.



\section{Additional visualization of false positives}
\label{Sec:Additional visualization of false positives}
In the main paper, we provide some visualizations of MI false positives. In this Supp., we provide more extensive visualizations of MI false positives in Fig.~\ref{fig:fp-maxvit-ppa}, \ref{fig:fp-d121-ppa}, \ref{fig:fp-r101-ppa}, \ref{fig:fp-r152-ppa}, \ref{fig:fp-r18-ifgmi}.

These false positive MI do not capture the visual identity features of the target individual in private training data, but are still considered successful attacks according to $\mathcal{F}_{Curr}$ with high confidence.

\section{Limitation}

While this study provides valuable insights into the limitations of the MI evaluation framework and propose a more reliable automated MI evaluation framework for future MI study, it is important to acknowledge certain limitations. One such limitation is the focus on specific architectures and datasets. While we strictly follow previous works \citep{zhang2020secret,chen2021knowledge,nguyen_2023_CVPR,struppek2022plug,qiu2024closer,koh2024vulnerability,ho2024model} to includes 26 MI setups, these setups may not include the latest architectures or dataset that are not considered in prevalent MI setups. Future research could expand upon our findings by exploring a wider range of model architectures and datasets. This would further shed the light of MI evaluation and contribute to the development of better MI evaluation frameworks.

\section{Ethical Statement}

This study examines the limitations of widely used evaluation frameworks for Model Inversion (MI) attacks, which hold critical implications for privacy and data security. Our analysis reveals an overestimation of MI attack success rates, underscoring the need for accurate and reliable evaluation metrics to avoid inflated perceptions of privacy risks. To support the research community, we propose a more reliable and cost-efficient MI evaluation framework based on MLLM. Furthermore, we release the code and a large-scale collection of MI reconstructed images upon publication, advocating for their ethical use to advance privacy protection.

\section{LLM Usage}

We used a large language model to help polish the grammar, wording, and other minor text issues in this manuscript. The authors are fully responsible for the ideas, analysis and conclusions in this submission.




\end{document}